\documentclass{article}

\usepackage{microtype}
\usepackage{graphicx}
\usepackage{subcaption}
\usepackage{booktabs}
\usepackage{amsmath,amssymb}
\usepackage[table,x11names]{xcolor}
\usepackage{soul}
\usepackage{bm}
\usepackage{hyperref}
\usepackage{makecell}
\usepackage{mathtools}
\usepackage{amsthm}
\usepackage{multicol}
\usepackage{multirow}
\usepackage{siunitx}
\usepackage{algorithm}
\usepackage{algpseudocode}
\usepackage{tikz}

\usepackage{arxiv}

\usepackage[utf8]{inputenc} 
\usepackage[T1]{fontenc}    
\usepackage{hyperref}       
\usepackage{url}            
\usepackage{booktabs}       
\usepackage{amsfonts}       
\usepackage{nicefrac}       
\usepackage{microtype}      
\usepackage{lipsum}		
\usepackage{graphicx}
\usepackage{natbib}
\usepackage{doi}

\title{A Physics-Aware Framework for Short-Term GPU Power Forecasting of AI Data Centers}

\author{Mohammad AlShaikh Saleh$^{1}$, Sanjay Chawla$^{2}$, Sertac Bayhan$^{3,4}$, Haitham Abu-Rub$^{1,3}$, and Ali Ghrayeb$^{1}$\\
$^{1}$College of Science and Engineering, Hamad Bin Khalifa University, Doha, Qatar\\
$^{2}$Qatar Computing Research Institute, Hamad Bin Khalifa University, Doha, Qatar\\
$^{3}$Qatar Environment and Energy Research Institute, Hamad Bin Khalifa University, Doha, Qatar\\
$^{4}$Department of Electrical \& Electronic Engineering, Gazi University, Ankara, Turkey}

\date{}

\hypersetup{
pdftitle={A template for the arxiv style},
pdfsubject={q-bio.NC, q-bio.QM},
pdfauthor={Mohammad AlShaikh Saleh, Sanjay Chawla, Sertac Bayhan, Haitham Abu-Rub, Ali Ghrayeb},
pdfkeywords={First keyword, Second keyword, More},
}

\begin{document}
\maketitle

\begin{figure*}[!htbp]
    \centering
    \vspace{-3.3\baselineskip}

    \begin{minipage}[t]{0.32\textwidth}
        \centering
        \includegraphics[width=\linewidth]{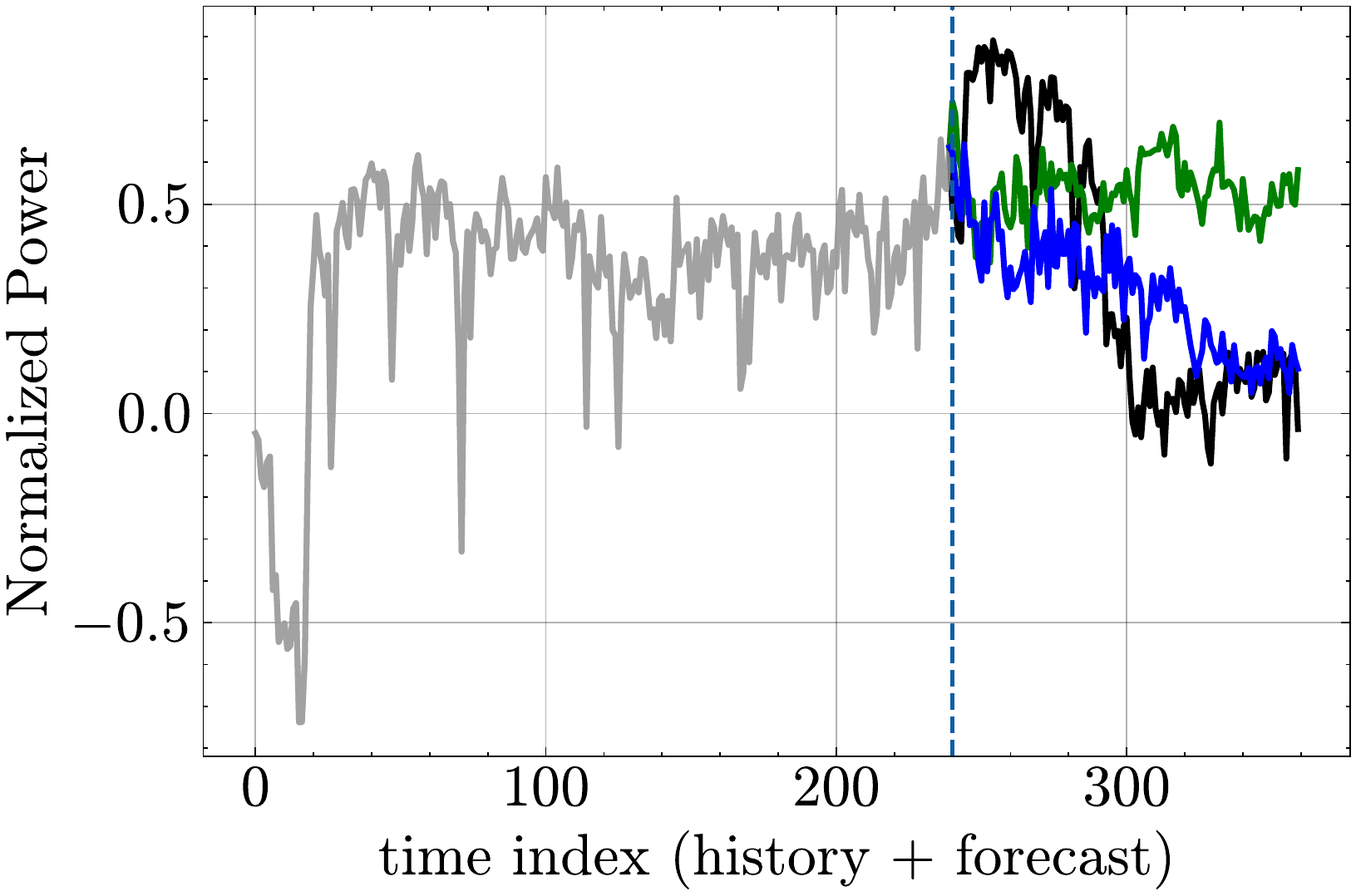}

        \vspace{0.3em}
        {\footnotesize\textbf{(a) Throttle-aware forecasting.}
        PI-DLinear captures GPU power throttling after the forecasting boundary with lower error than DLinear
        (MAE/RMSE: 0.4454/0.5103 vs.\ 0.4703/0.5528).}
    \end{minipage}\hfill
    \begin{minipage}[t]{0.32\textwidth}
        \centering
        \includegraphics[width=\linewidth]{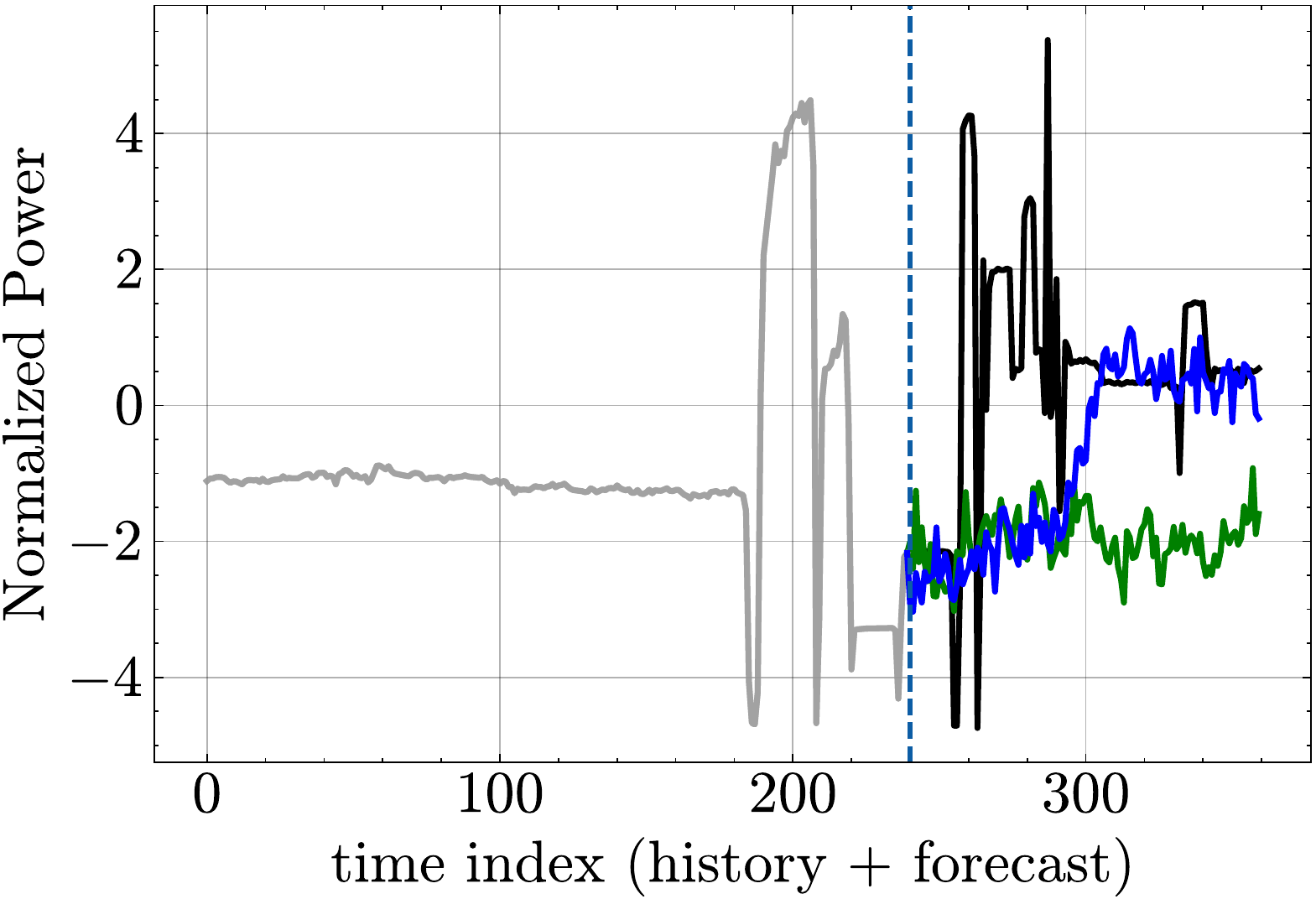}

        \vspace{0.3em}
        {\footnotesize\textbf{(b) Transient recovery performance.}
        Under abrupt load fluctuations, PI-DLinear recovers forecasting accuracy more robustly than DLinear
        (MAE/RMSE: 1.4650/2.3061 vs.\ 2.5082/2.8610).}
    \end{minipage}\hfill
    \begin{minipage}[t]{0.32\textwidth}
        \centering
        \includegraphics[width=\linewidth]{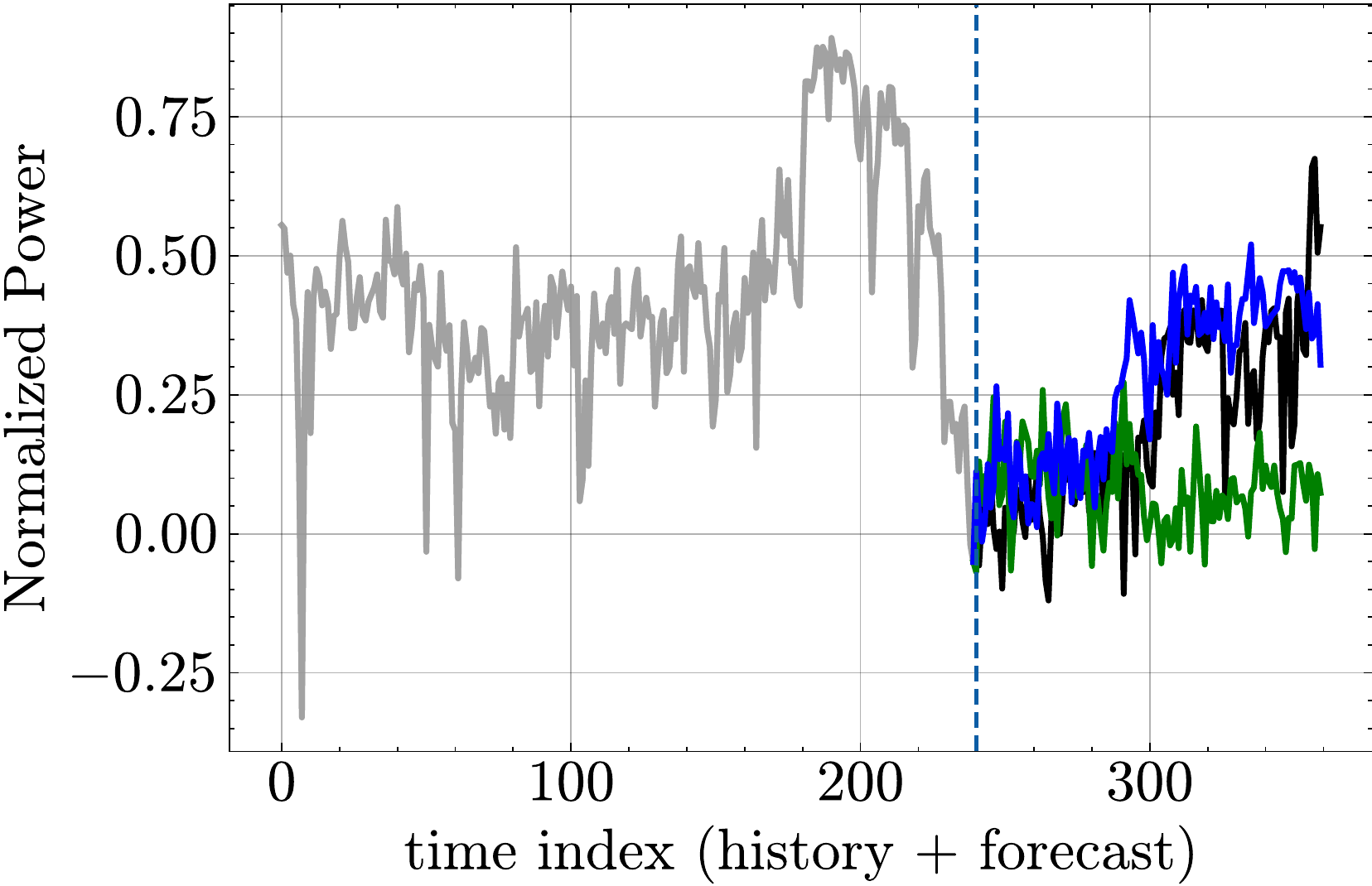}

        \vspace{0.3em}
        {\footnotesize\textbf{(c) Post-throttle stability.}
        After throttling subsides, PI-DLinear maintains stable predictions with lower error than DLinear
        (MAE/RMSE: 0.1112/0.1469 vs.\ 0.1795/0.2274).}
    \end{minipage}

    \vspace{0.4em}

    \begin{minipage}[t]{\textwidth}
        \centering
        \includegraphics[width=0.50\textwidth]{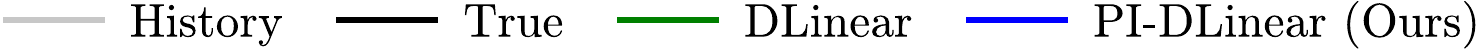}
    \end{minipage}

    \caption{\textbf{Comparison of DLinear and PI-DLinear across power throttling, transient recovery, and post-event stability regimes.}
    PI-DLinear consistently achieves lower prediction error, enabling accurate throttling characterization, faster recovery from sudden AI load changes, and stable forecasting behavior.}
    \label{fig:teaser}

    \vspace{15pt}
\end{figure*}


\begin{abstract}
AI data centers experience rapid fluctuations in power demand due
to the heterogeneity of computational tasks that they have to support. For example,
the power profile of inference and training of large language models (LLMs) is quite distinct and big
divergences can result in the instability of the underlying electricity grid. In this
paper we propose, to the best of our knowledge, the first physics-informed
DLinear time-series model that can accurately forecast power utilization
of an AI data center 5-80 minutes (short-term forecasting) into the future. The physics, based on a multi-node lumped thermal resistance-capacitance (RC) network consistent with Newton’s law of cooling, is captured using newly derived time-dependent ordinary differential equations (ODE) that separately models and interlinks power consumption with the GPU compute and memory utilization and temperature. The resulting model, that we refer to as \textbf{PI-DLinear}, trained and evaluated on a real AI data center dataset and is not only more accurate than the state-of-the-art (SOTA) models tested, but the forecast profile respects the underlying physics under power throttling and load transient events. Relative to the SOTA transformer-based and non-transformer-based models, improvements in forecasting accuracy (averaged across all look-back and prediction windows) range from 0.782\%--39.08\% for MSE, 0.993\%--51.82\% for MAE, and 0.370\%--22.28\% for RMSE.
\end{abstract}
\vspace{-0.35\baselineskip}
\textbf{Keywords}\hspace{0.3cm} AI data centers, DLinear, energy-efficient computing, Physics-aware modeling, power forecasting, power grid resiliency, RC thermal modelling, and transformers.

\section{Introduction}
\label{intro}
\subsection{Problem Statement}
The data center industry is rapidly growing, fueled by the rising demand for cloud services, advancements in Artificial Intelligence (AI) and Machine Learning (ML), and the need for data storage. This surge is anticipated to significantly increase global electricity demand related to AI, influenced by a variety of technological, economic, and societal factors \cite{lin2023adapting}. This decade has witnessed substantial investments from major tech hyperscalers in expanding data center infrastructures worldwide \cite{google2024datacenters,oracle2024datacenters}. However, these escalating demands present challenges for power grids, raising concerns about their ability to accommodate high-density power loads, as noted in PJM's report on overloaded lines and increasing electricity costs \cite{aurora2024pjm}. Environmental pressures are also prompting data centers to adopt modular designs and renewable energy sources. Moreover, AI-driven data centers are experiencing unprecedented power densities per rack, leading to significant transitory power fluctuations similar to those caused by Electric Vehicles (EVs) and renewable energy systems \cite{lin2024exploding}. 

Recently, with large data centers consuming tens to hundreds of megawatts, power changes of several megawatts can occur within mere seconds, potentially disrupting grid frequency control and necessitating faster frequency regulation responses. Notably, modern AI workloads demand much higher power densities, ranging from 300\,\unit{\watt} to 1,200\,\unit{\watt} per GPU, and they exhibit rapid power fluctuations, such as exceeding 132\,\unit{\kilo\watt/\second} at the rack level with NVIDIA's GB200 NVL72 \cite{schneidernvidiaaireference}. In light of this, one of the crucial ways to enhance grid reliability is to forecast data center power consumption to inform control strategies such as primary control, secondary control, tertiary control, and Automatic Generation Control (AGC), especially when grappling with substantial power transients driven by AI loads, harmonics, and frequency fluctuations \cite{li2024llmtransients}. Additionally, predicting upcoming voltage sags and swells in power consumption can aid AGC in effectively preparing for rapid ramp-up events.

Therefore, accurate short-term forecasts of AI data-center demand help grid operators schedule balancing actions and reserve requirements and make real-time network operational decisions more efficient, improving reliability and lowering operating costs. From a grid-planning perspective, short-term forecasts also make it easier to treat data centers as flexible loads (e.g., shifting or curtailing demand through demand response), which can reduce peak stress, avoid/defer costly grid upgrades, and reduce the gap between generation capacity and load demand. Thus, power forecasting continues to be a significant area of research, contributing directly to both improved energy management and system optimization.

In recent times, transformer-based (deep learning) models have revolutionized time series forecasting by effectively capturing both short-term and long-term dependencies. To evaluate these advanced models, the Time Series Library (TSLib) \cite{wu2023timesnettemporal2dvariationmodeling} was used, a comprehensive library featuring state-of-the-art (SOTA) models, most of which are transformer-based. Nevertheless, the problem with deep learning models in general is that they can lead to physically inconsistent predictions and are less reliable outside the range (out-of-distribution) of the training data, especially in the event of power throttling, abrupt load fluctuations and AI workloads, and post-throttle stability. These are all cases that our proposed model, which we call physics-informed-DLinear (\textbf{PI-DLinear}), addresses as observed in Fig. \ref{fig:teaser}. Therefore, our paper integrates the first physics-informed framework that guides the model with scientific principles, improving generalization, interpretability, and robustness by anchoring the learning to real physical mechanisms rather than just patterns in the training data for power forecasting across diverse AI workloads (LLMs, vision networks, and GNNs) and multiple temporal historical and prediction scales. The proposed model was tested on the MIT supercloud dataset \cite{samsi2021mitsuperclouddataset}, which, to the best of the authors' knowledge, is the only publicly available dataset demonstrating the impact of AI workloads on data center power consumption. 

Physics-informed neural networks (PINNs), a field also referred to as scientific machine learning, integrate data-driven and physics-driven approaches to address challenges in science and engineering \cite{raissi2018deep,MCCLENNY2023111722}. These algorithms leverage substantial prior knowledge in the form of algebraic and/or differential equations describing the relationships among variables, making them particularly well-suited for power forecasting applications. However, the problem is that no time-dependent ODE/PDE was available that takes into account the interlink of power consumption with GPU temperature and utilization and memory temperature and utilization, which is a prerequisite to construct a physics-aware framework. Consequently, this is also the first work that required to derive new ODEs using a coupled two-node resistance-capacitance (RC) thermal network to enable their smooth integration as physics-informed constraints for power forecasting in AI data centers.

\subsection{Contributions}
To this end, the main contributions are as follows:
\begin{itemize}
    \item We present the first Physics-Informed DLinear (PI-DLinear) model that incorporates a multi-node lumped thermal RC network, consistent with Newton's law of cooling, for GPU power forecasting across diverse AI workloads (LLMs, vision networks, and GNNs) along with a systematic evaluation across multiple temporal scales, with look-back windows of 240, 360, 480, and 600 minutes and prediction horizons of 5, 10, 20, 40, and 80 minutes ahead.
    \item To the best of our knowledge, this is the first work that needed to derive new ODEs using the coupled two-node RC thermal network to enable their smooth integration as physics-informed constraints for power forecasting in AI data centers, especially given the limited number of high-importance features in the dataset used. Specifically, the ODEs relate the GPU power to GPU temperature and the memory power to memory temperature, allowing for the modeling of GPU and memory as coupled thermal nodes. 
    \item Detecting and predicting power throttling using the proposed PI-DLinear was achieved, delivering improved transient recovery forecasting performance under abrupt AI load fluctuations and stable predictions even after power throttling subsides.
    \item A comprehensive comparison against SOTA transformer-based and non-transformer-based forecasting models demonstrated improvements in prediction accuracy (averaged across all look-back and prediction windows) ranging from 0.782\%--39.08\% for MSE, 0.993\%--51.82\% for MAE, and 0.370\%--22.28\% for RMSE.
    \item The code is provided, allowing researchers to replicate, use, and extend the work for further advancements in power forecasting of AI data centers.  
\end{itemize}
\subsection{Related Works}
Current research \cite{amvrosiadis2018diversity, blocher2021switches, wilkins2024hybrid} explores various energy-related factors such as workloads and failure rates. However, even diverse datasets with unknown job profiles complicate prediction efforts. As noted in \cite{wang2024utilization}, there are opportunities in reinforcement learning that could enhance scheduling efficiency. Energy management for GPUs is addressed in \cite{rossi2015holtwinters,meisner2009powernap}, focusing on optimizing power based on active or idle states and predictive modeling. Moreover, regression techniques, including Auto Regressive Integrated Moving Average (ARIMA) and fault tree methodologies, are utilized in \cite{shoukourian2020lstm} to forecast power consumption and failure incidents in data center facilities. Furthermore, \cite{wilkins2024hybrid} investigates how user behavior influences energy usage, while \cite{bai2022dnnabacus} showcases a convolutional neural network (CNN) approach that improves predictions of GPU power consumption for large language models (LLMs) over traditional methods like ARIMA. Additionally, a Deep Neural Network (DNN) is introduced in \cite{patel2024llmcloud} to estimate the computational costs of LLM training in cloud environments. Given the dynamic demands of AI and high-performance computing workloads, accurate short-term power forecasting in data centers is crucial \cite{hu2021gpuworkloads}. Traditional models like ARIMA often fall short in recognizing complex patterns within high-dimensional datasets, while LSTM, GRU, and CNN models demonstrated strong performance in handling these challenges, particularly on the MIT dataset \cite{mughees2025short}. 

Deep learning techniques stand out for their ability to decipher intricate data patterns, offering promising solutions to the limitations mentioned above \cite{9917400}. In particular, recurrent neural networks (RNNs) and their long short-term memory (LSTM) variants have garnered significant attention for their effectiveness in time-series forecasting \cite{9496627}. LSTMs adeptly tackle the vanishing and exploding gradient issues typical of RNNs, enabling them to capture long-range dependencies in time-series data \cite{9200614}. Furthermore, bidirectional LSTM (BiLSTM) networks enhance predictive accuracy by leveraging both past and future contextual information \cite{9889183}.

However, recently, transformers have emerged as the leading architecture for sequence modeling, showcasing exceptional performance across a range of applications, including natural language processing (NLP), speech recognition, and computer vision. Recently, their use for analyzing time-series data has gained traction, as discussed in \cite{wen2022transformers}. Key transformer-based models for time-series forecasting include LogTrans \cite{li2019enhancing}, Informer \cite{zhou2023beyond}, Autoformer \cite{wu2021autoformer}, Pyraformer \cite{liu2022pyraformer}, Triformer \cite{cirstea2022triformer}, and FEDformer \cite{zhou2022fedformer}. 

The main working power of transformers-based forecasting models is from their multi-head self-attention mechanism, which has a remarkable capability of extracting semantic correlations among elements in a long sequence. Conventional transformer-based time-series forecasting models predominantly assume numeric, particularly continuous, inputs and therefore often overlook recurring motifs that are fundamental to many real-world temporal dynamics. In addition, the widespread strategy of segmenting long sequences into fixed-size sliding windows for supervised training restricts the model’s ability to exploit global contextual information beyond the current segment.


\section{Proposed Forecasting Framework}
\label{method_sec}
\subsection{Task Definition: Short-Term Power Forecasting}
We study short-term forecasting of data-center electricity load under AI workloads (e.g., inference bursts and training interruptions). For multivariate time series forecasting, given historical data
$\mathcal{X} = \{ \mathbf{x}_1, \mathbf{x}_2, \dots, \mathbf{x}_L \} \in \mathbb{R}^{L \times C}$,
where $L$ denotes the look-back window length and $C$ is the number of co-variates,
$\mathbf{x}_t = [x_t^1, x_t^2, \dots, x_t^C]$ represents the multivariate observation at time step $t$, which includes hardware utilization signals and thermal sensors (GPU utilization, memory utilization, GPU temperature, memory temperature) and the power draw.
The objective is to predict the future power consumption sequence
$\mathbf{y} = \{ P_{L+1}, P_{L+2}, \dots, P_{L+T} \} \in \mathbb{R}^{T}$
over a forecasting horizon of length $T$, where $P_t$
denotes the power consumption at time step $t$. Overall, the forecasting task can be formulated as a mapping
$\mathbb{R}^{L \times C} \rightarrow \mathbb{R}^{T}$, where the input consists of $C>1$ variables over $L$ time steps and the output is the power forecast over $T$ future time steps. It is worth noting that the input variables are commonly treated as multiple channels.

Our forecasting target is the power consumption $P_t$ (in Watts). Given a look-back window of length $L$ and a forecasting horizon $T$, we define
\begin{equation}
\begin{split}
&\mathbf{X}_{t-L+1:t} = [\mathbf{x}_{t-L+1},\dots,\mathbf{x}_{t}] \in \mathbb{R}^{L\times C},\\
\qquad
&\mathbf{y}_{t+1:t+T} = [P_{t+1},P_{t+2},\dots,P_{t+T}]^\top \in \mathbb{R}^{T}.
\end{split}
\end{equation}
We learn a forecasting model $f_{\theta}$ (Transformer or non-Transformer) that predicts future power, written as
\begin{equation}
\widehat{\mathbf{y}}_{t+1:t+T} = f_{\theta}\!\left(\mathbf{X}_{t-L+1:t}\right) \in \mathbb{R}^{T}.
\end{equation}
The standard supervised objective minimizes a prediction loss $\mathcal{L}_{\mathrm{pred}}$ over $N$ training samples as
\begin{equation}
\begin{split}
&\min_{\theta}\;\; \frac{1}{N}\sum_{n=1}^{
N}\mathcal{L}_{\mathrm{pred}}\!\left(\widehat{\mathbf{y}}^{(n)},\mathbf{y}^{(n)}\right),\\
\quad
&\mathcal{L}_{\mathrm{pred}}(\widehat{\mathbf{y}},\mathbf{y})=\frac{1}{T}\sum_{k=1}^{T}\left(\widehat{P}_{t+k}-P_{t+k}\right)^2,
\end{split}
\end{equation}
where $\widehat{P}_{t+k}$ denotes the $k$-step-ahead predicted power and $\theta$ is the learnable parameter that includes the weights and biases of the neural network.

\subsection{DLinear}
\begin{figure*}[!t]
    \centering
    \includegraphics[width=1\linewidth]{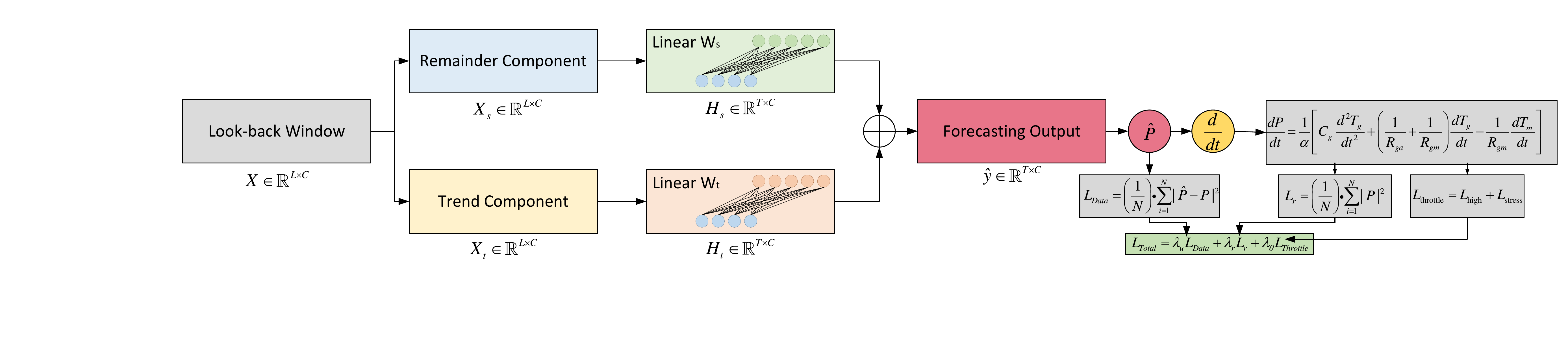}
    \caption{\textbf{PI-DLinear Architecture.} The base DLinear model (top) decomposes the input look-back window $\mathbf{X} \in \mathbb{R}^{L \times C}$ into seasonal/remainder ($\mathbf{X}_s$) and trend ($\mathbf{X}_t$) components, which are independently projected to the forecast horizon via linear layers $\mathbf{H}_s$ and $\mathbf{H}_t$, then summed to produce the full multivariate forecast $\widehat{\mathbf{Y}} \in \mathbb{R}^{H\times C}$, from which the power channel $\widehat{\mathbf{y}} \in \mathbb{R}^{H}$ is extracted. Our physics-informed extension introduces three loss components: (1) $\mathcal{L}_{\text{data}}$: standard MSE loss with ground truth, (2) $\mathcal{L}_{r}$: coupled RC thermal network residual enforcing consistency between predicted power and observed temperatures $T^g$, $T^m$, and (3) $\mathcal{L}_{\text{throttle}}$: utilization-based constraint encoding that high GPU/memory utilization should precede power reduction. The RC parameters for the aggregate GPU rack ($C_g=5.408\times10^6$\,\unit{\joule/\kelvin}, $C_m=5.481\times 10^6$\,\unit{\joule/\kelvin}, $R_{ga}=2.037\times 10^{-3}$\,\unit{\kelvin/\watt}, $R_{ma}=2.055\times 10^{-3}$\,\unit{\kelvin/\watt}, $R_{gm}=6.064\times 10^{-4}$\,\unit{\kelvin/\watt}, $T_a=27^\circ$\unit{C}, $\alpha=0.5085$) are pre-estimated via RLS using the training data and then are kept fixed for validation and testing.}
    \label{fig:pidlinear_architecture}
\end{figure*}

DLinear combines a decomposition approach inspired by Autoformer and FEDformer with linear layers \cite{zeng2023transformers}. It begins by decomposing the historical time series data into a trend component via a moving average kernel and a seasonal/remainder component. Then, two separate one-layer linear layers are applied to each component, and their outputs are summed to yield the final prediction as
\begin{equation}
    H_s=W_sX_s\in \mathbb{R}^{T\times C}, \hspace{0.5cm} W_s\in \mathbb{R}^{T\times L},
\end{equation}
\begin{equation}
    H_t=W_tX_t\in \mathbb{R}^{T\times C}, \hspace{0.5cm} W_t\in \mathbb{R}^{T\times L},
\end{equation}
\begin{equation}
    \widehat{\mathbf{y}}=H_s+H_t,
\end{equation}
where $H_s$ and $H_t$ denote the output values of the single-layer linear networks corresponding to the residual and trend components, respectively. $W_s$ and $W_t$ symbolize the single-layer linear networks associated with the residual and trend components, as further shown in Fig. \ref{fig:pidlinear_architecture}. It is worth noting that if the dataset's variables have different characteristics, i.e., different seasonality and trends, then sharing weights across the co-variates may not perform as desired. Hence, by explicitly addressing the trend, DLinear enhances the basic linear model's performance, especially in datasets exhibiting clear trends \cite{zeng2023transformers}.

\subsection{Physics-Aware Regularization via a Compact Thermal RC Model}
\subsubsection{Derived ODEs using the Thermal Circuit RC Model}
Power consumption and temperature evolution are coupled through energy conservation and Newtonian cooling. Following compact thermal modeling, we adopt a lumped-parameter RC network with two coupled thermal states: GPU temperature $T_g(t)$ and memory temperature $T_m(t)$. We also have $T_a$ denoting the effective ambient/sink temperature. In this paper, the lumped-parameter model is used to simplify a spatially distributed system by representing it as discrete nodes with uniform properties, where each node (e.g., GPU, Memory) has a single temperature value rather than a continuous temperature field. Instead of solving partial differential equations (PDEs) that describe how temperature varies across every point in space, we derive ODEs that describe how the temperature of each lumped node evolves over time. The lumped assumption can be made here because GPUs are designed with high thermal conductivity (copper heat spreaders, thermal paste), so the temperature distribution is relatively uniform within each component, validating the lumped-parameter approximation approach.

Applying Kirchhoff's Current Law at each thermal node (see \ref{APP:Circuit} for the full derivation), the energy-balance ODEs for the GPU and memory nodes are
\begin{equation}
C_g \frac{dT_g}{dt} = \alpha\,P - \frac{T_g - T_a}{R_{ga}} - \frac{T_g - T_m}{R_{gm}},
\label{eq:gpu_ode}
\end{equation}
\begin{equation}
C_m \frac{dT_m}{dt} = (1-\alpha)\,P - \frac{T_m - T_a}{R_{ma}} + \frac{T_g - T_m}{R_{gm}},
\label{eq:mem_ode}
\end{equation}
where $C_g, C_m > 0$ are the thermal capacitances (heat storage) of the GPU and memory respectively, $R_{ga}, R_{ma} > 0$ are the thermal resistances from each component to the ambient, $R_{gm} > 0$ is the thermal coupling resistance between the GPU and memory, and $\alpha \in [0,1]$ is a latent power split parameter defined below. The terms $(T_g - T_a)/R_{ga}$ and $(T_m - T_a)/R_{ma}$ correspond to Newton's law of cooling in lumped form (i.e., heat loss proportional to temperature difference).

Solving Equation~\eqref{eq:gpu_ode} for $P$ and differentiating with respect to time yields the following \textbf{power rate constraint}:
\begin{equation}
\frac{dP}{dt} = \frac{1}{\alpha}\left[ C_g \frac{d^2T_g}{dt^2} + \frac{1}{R_{ga}}\frac{dT_g}{dt} + \frac{1}{R_{gm}}\left(\frac{dT_g}{dt} - \frac{dT_m}{dt}\right) \right],
\label{eq:dpdt_constraint1}
\end{equation}
where the memory temperature derivative $dT_m/dt$ is itself governed by Equation~\eqref{eq:mem_ode}, thereby coupling both thermal nodes into the constraint. The thermal capacitance and resistance terms are empirically determined using recursive least squares (RLS), a common practice in thermal RC modeling. We acquire the derivatives of the proposed PI-DLinear model (surrogate model) with respect to time $t$ through chain rule and differentiating function compositions by automatic differentiation ($\mathrm{AutoDiff}$) \cite{raissi2018deep,braga2024physics}. It is worth noting that GPUs convert nearly 99\% of electrical energy into heat through transistor switching losses, resistive losses in interconnects, and leakage currents. Also, since the ambient temperature $T_a$ was not provided, it was assumed that due to cooling, $T_a$ was kept constant at room temperature (the minimum observed temperature, which was $\approx 27^\circ \mathrm{C}$).

Here, we observe only the total power $P(t)$ rather than its partition into GPU and memory contributions. We therefore introduce a latent split parameter $\alpha\in[0,1]$ as
\begin{equation}
P_g(t) = \alpha\,P(t),
\qquad
P_m(t) = (1-\alpha)\,P(t).
\label{eq:power_split}
\end{equation}
In practice, $\alpha$ can be treated as a learned scalar or a learned function of workload features.

\subsubsection{Power Throttling Constraint}
GPU power throttling is an automatic protection mechanism that reduces power consumption to prevent thermal damage when the processor approaches critical operating limits. In light of this, we observe from the MIT Supercloud dataset that throttling events, characterized by sudden power drops exceeding 15\%, are strongly correlated with sustained high utilization rather than extreme temperatures alone, as the critical GPU temperatures (see Fig.~\ref{fig:Powers}) never reach the limits $80^\circ\mathrm{C}-90^\circ\mathrm{C}$ outlined in the product brief \cite{nvidia2018teslaV100PCIe}. This physical insight motivates our throttling-aware loss function, which ensures that when utilization exceeds threshold $\theta_U\approx 90\%$, power should not increase, i.e., $U_t > \theta_U \Rightarrow \Delta \hat{P}_t \leq 0$, and simultaneously high temperature and high utilization strongly indicate imminent throttling, i.e., $(U_t > \theta_U) \land (T^g_t > \theta_{T}) \Rightarrow \Delta \hat{P}_t \leq 0$. We also set $\theta_{T}$ to be at the $95^{\mathrm{th}}$ percentile, even though we don't reach the $80^\circ\mathrm{C}-90^\circ\mathrm{C}$ limits, yet, the rise in the temperature is highly correlated with the rise in the utilization. These constraints are combined into a single differentiable loss as
\begin{equation}
\mathcal{L}_{\text{throttle}} = \mathcal{L}_{\text{high}}+\mathcal{L}_{\text{stress}},
\label{eq:throttle_loss}
\end{equation}
where each component is penalized, given that it violates the corresponding physical constraint set. Therefore, we have
\begin{equation}
\begin{split}
&\mathcal{L}_{\text{high}} = \frac{1}{H-1} \sum_{t:\, U_t > \theta_U} \max(0,\, \Delta \hat{P}_t)^2, \\
&\mathcal{L}_{\text{stress}} = \frac{1}{H-1} \sum_{t:\, U_t > \theta_U,\, T^g_t > \theta_{T}} \max(0,\, \Delta \hat{P}_t)^2,
\end{split}
\label{eq:throttle_components}
\end{equation}
where $H$ is the prediction horizon, $\Delta \hat{P}_t = \hat{P}_{t+1} - \hat{P}_t$ is the predicted power change, $U_t = \alpha \cdot u_t^{(g)} + (1-\alpha) \cdot u_t^{(m)}$ is the weighted GPU-memory utilization, and $\theta_U$, $\theta_{T}$ are the utilization and temperature thresholds respectively.

\subsubsection{Loss Function Formulation} 
The weighting parameter $\lambda$, which controls the trade-off between data loss and physics loss along with the hyperparameters of the PINN model, is optimized by obtaining a minimum of the subsequent weighted loss function \cite{braga2020fundamentals}
\begin{equation}
    \mathcal{L}=\lambda_u\cdot \mathcal{L}_{\mathrm{Data}}+\lambda_r\cdot \mathcal{L}_r+\lambda_\theta\cdot \mathcal{L}_{\mathrm{throttle}},
    \label{eq:lossTot}
\end{equation}
through different optimization methods like stochastic gradient descent, Adam, adaptive gradient descent, RMS Prop, etc. \cite{braga2020fundamentals}. The weights $\lambda_u$ (data weight), $\lambda_r$ (residual weight), $\lambda_\theta$ (throttle weight) are carefully selected to keep training balanced among the three losses. In this study, the weights are self-adaptive and are therefore updated by gradient ascent in log-space to keep them positive, i.e. $\eta_u=\log{\lambda_u}$, $\eta_r=\log{\lambda_r}$, $\eta_\theta=\log{\lambda_\theta}$ and $\eta\leftarrow\eta+\gamma\nabla_\eta\mathcal{L}$ while clipping $\eta$ to $[\lambda_{\mathrm{min}},\lambda_{\mathrm{max}}]$ to avoid instability and collapse to extremes. We also compute $\mathcal{L}_r$ during training merely on the sequence length using the observed $T_g$ and $T_m$ so they are rolled out using the learned RC model driven by the predicted future power $\hat{P}$.

In the proposed algorithm, the weighting factors are considered to be scalar, and thus, the PINNs weight loss function is adhered to in the training optimization process and can be written as
\begin{equation}
\begin{split}
&\mathcal{L}_{\mathrm{Data}}(\bm{\lambda}_u) \doteq \frac{1}{N_u} \sum_{i=1}^{N_u} \left|\hat{P_i}(\bm{x}_u^i, t_u^i; \Phi) - P_i \right|^2, \\
&\mathcal{L}_r(\bm{\lambda}_r) \doteq \frac{1}{N_r} \sum_{i=1}^{N_r} |P(\bm{x}_r^i, t_r^i; \Phi,\phi)|^2.
\end{split}
\label{eq:LMod}
\end{equation}

where $\lambda_u$, $\lambda_r$, and $\lambda_\theta$ are the weights for the data, physics (residual), and power throttle components, respectively, $\mathcal{L}_u$ is the data loss component at the collocation points, $\mathcal{L}_r$ is the physics (residual) loss component, $\mathcal{L}_\theta$ is the power throttling loss component $N_u, N_r,N_\theta$ are the collocation points, $\bm{x}^i$ represents the $i$-th collocation point of the co-variates in the 5-dimensional feature space, $t^i$ is the corresponding time coordinate feature,
$\hat{P}(\bm{x}_u^i, t_u^i; \Phi)$ is the DLinear prediction at the collocation points $(\bm{x}_P^i, t_P^i)$, and $u_i$ is the actual measurement in the dataset. 

\section{Experimental Results and Discussion}
\label{exp_sec}
This brings us to the following questions. \textit{1. Does adding a physics-based constraint to the loss function improve the power forecasting performance for AI-centric data centers?} \textit{2. What is the computational burden on running the proposed physics-aware model?} \textit{3. What impact does the physics weighting/regularization factor $\lambda$ have on the power forecasting performance?} \textit{4. Can PI-DLinear still perform in the event of power throttling?} 

\subsection{Data Collection and Pre-Processing}
The gathered data encompasses GPU power consumption, often the largest portion of total power used in AI workloads, along with metrics like memory utilization, GPU temperature, and storage. Following data collection, the raw data undergoes pre-processing steps, such as Min-Max normalization and data slicing, to prepare it for integration into the forecasting models.

In this study, we address the data center power consumption forecasting problem using a real-world dataset from the MIT Supercloud \cite{samsi2021mitsuperclouddataset}, a high-performance computing (HPC) system (GPU: Nvidia Volta V100, CPU: Intel Xeon Gold 6248). It is worth noting that, to the best of the authors' knowledge, this is the only publicly available dataset demonstrating the impact of AI workloads on data center power consumption. 

\begin{figure}[t]
\centering
\setlength{\abovecaptionskip}{2pt}
\setlength{\belowcaptionskip}{-6pt}

\begin{subfigure}{\linewidth}
    \centering
    \includegraphics[width=0.495\linewidth]{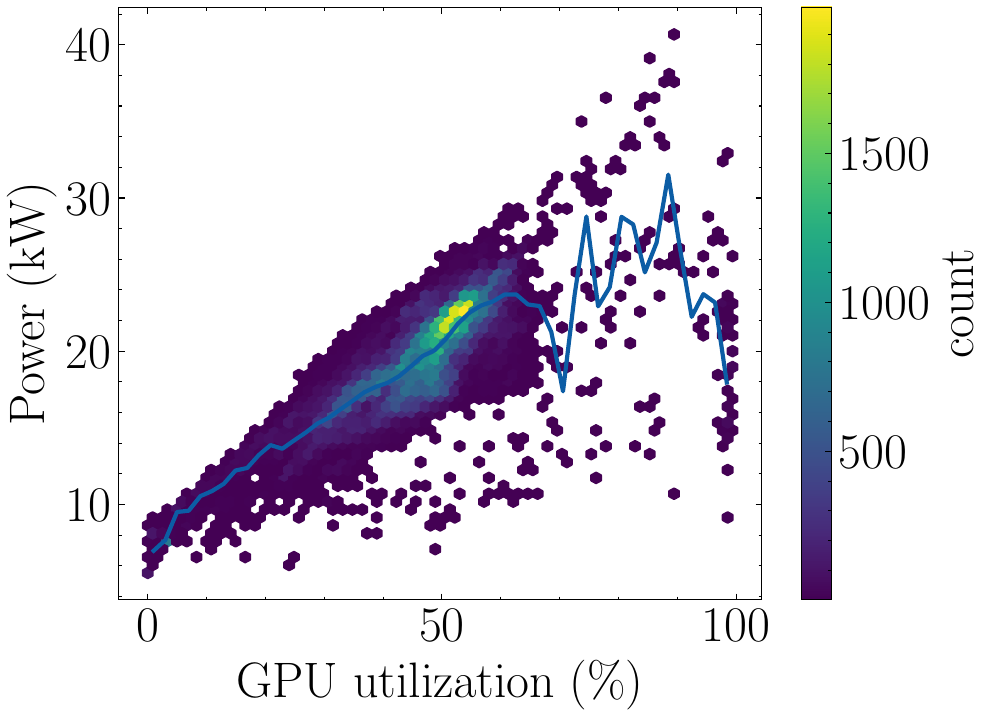}\hfill
    \includegraphics[width=0.495\linewidth]{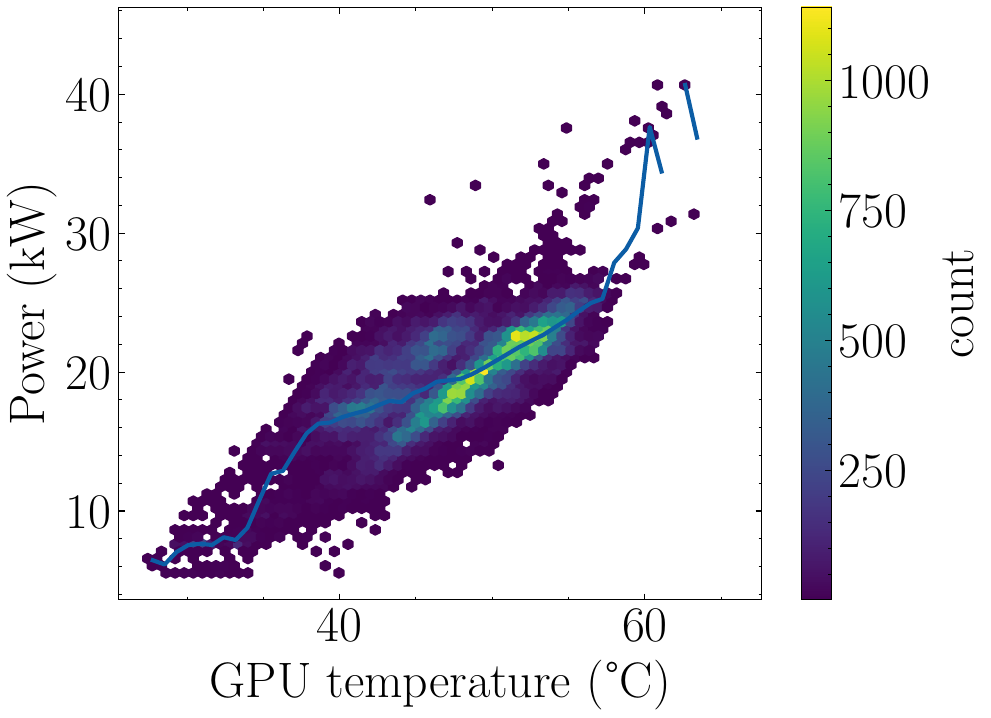}
    \caption{Power vs. utilization (left) and power vs. temperature (right).}
\end{subfigure}

\vspace{5pt}

\begin{subfigure}{\linewidth}
    \centering
    \includegraphics[width=0.50\linewidth]{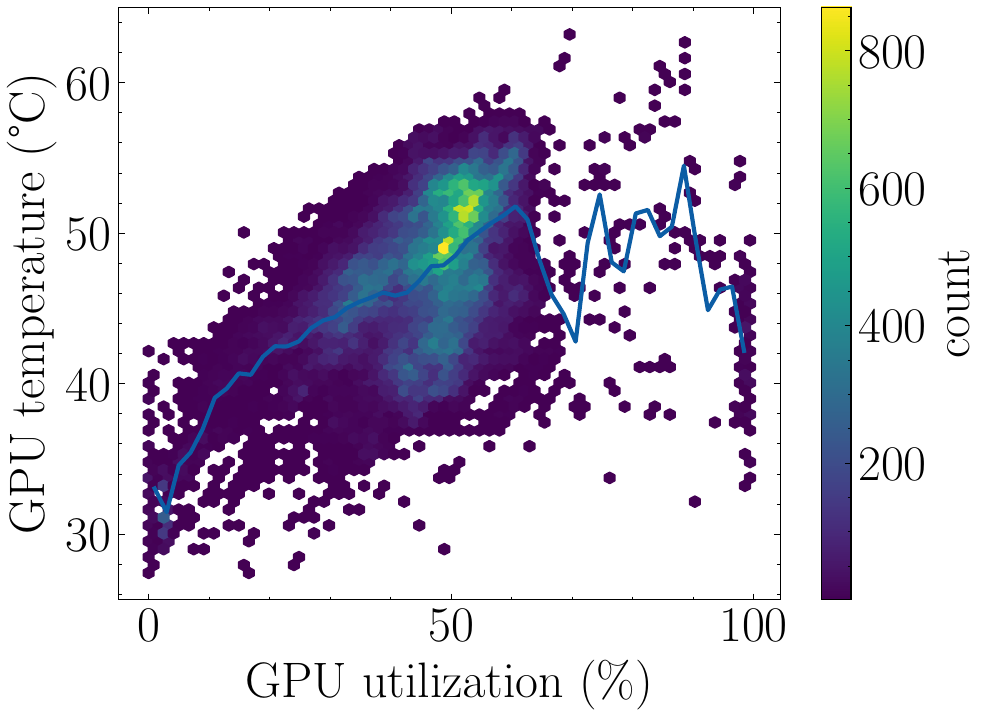}
    \caption{Temperature vs. utilization.}
\end{subfigure}
\vspace{0.5pt}
\caption{Relationships observed in the MIT Supercloud dataset: power vs.\ utilization (left), power vs.\ temperature (right), and temperature vs.\ utilization (bottom). Lighter regions indicate higher count, and darker regions indicate lower count.}
\label{fig:Powers}
\end{figure}

\begin{figure}[!b]
\vspace{-\baselineskip}
    \centering
    \includegraphics[width=0.75\linewidth]{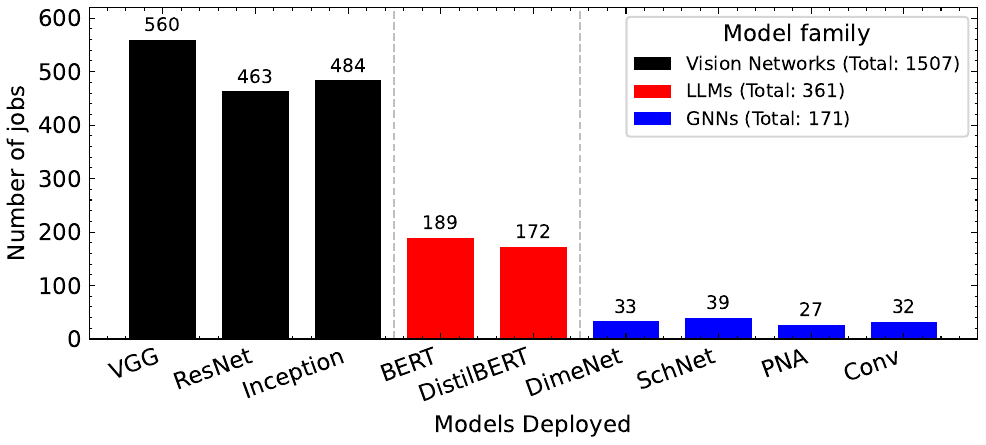}
    \caption{Computational job distribution across the AI workloads present in the MIT Supercloud dataset, namely, vision networks, large language models (LLMs), and graph neural networks (GNNs).}
    \label{fig:jobs}
\end{figure}

The dataset spans February to October 2021 and includes 100-millisecond interval logs of GPU/CPU utilization, scheduling details, and physical critical parameters like power draw, temperature, and utilization plotted in Fig. \ref{fig:Powers}. Key GPU metrics include power, memory, utilization, and temperature, with anonymized user data organized by job ID and node. Aggregated GPU power consumption peaks at 45 kW across 448 GPUs. The dataset details workload composition, dominated by vision networks (e.g., U-Net: 1,431 jobs; VGG, ResNet, and Inception follow), language models (e.g., BERT: 189 jobs; DistillBERT: 172 jobs), and graph neural networks (SchNet, DimeNetm, PNA, and conv) as shown in Fig. \ref{fig:jobs}. Pre-processing maintains a 1-minute granularity, with power consumption aggregated by job ID and node to reflect total power drawn from the local distribution system. After normalization via a min-max scalar, the data uses different look-back windows to predict different prediction lengths in minutes ahead. 

\begin{table}[t]
\centering
\footnotesize
\caption{Summary of the MIT Superclouddataset used in this study.}
\label{tab:dataset_summary}
\renewcommand{\arraystretch}{1.15}
\setlength{\tabcolsep}{9pt}
\begin{tabular}{l c}
\toprule
\textbf{Attribute} & \textbf{MIT Supercloud GPU Trace} \\
\midrule
Co-variates &
\makecell{7 (GPU utilization, memory utilization,\\
free memory, used memory,\\
GPU temperature, memory temperature,\\
power draw)} \\
\# Timesteps  & $\sim$330{,}500 \\
Granularity  & 1\,\unit{\minute} \\
Duration     & $\sim$238 days \\
\bottomrule
\end{tabular}
\end{table}
This dataset summary (Table \ref{tab:dataset_summary}) indicates that the experiments are conducted on a high-resolution, long-duration multivariate operational trace from the MIT Supercloud GPU cluster. After pre-processing the data, seven variates were sampled at 1-minute granularity, yielding approximately 330k timesteps spanning about 238 days, which is well-suited for evaluating both short-horizon and longer-horizon forecasting under realistic AI workload variability. At the end, 5 co-variates were chosen (omitting memory used and memory free due to low observed correlation), namely the memory utilization (in \%), memory temperature, GPU utilization (in \%), GPU temperature, and the power drawn. These features were all used for the proposed model along with the SOTA benchmark models.

\subsection{Model Training}
In time-series forecasting, various deep learning architectures can be utilized, but this paper focuses on 16 state-of-the-art models renowned for their effectiveness with sequential data, namely, Transformer, iTransformer, TimeXer, TiDE, TSMixer, Reformer, PatchTST, Nonstationary transformer, LightTS, FiLM, FEDformer, Pyraformer, DLinear, Crossformer, NLinear, and Linear. Each model is trained, its hyperparameters are fine-tuned, and validated for performance evaluation. The dataset was preprocessed to ensure consistent formatting, remove missing or corrupted entries, and choose the high-importance co-variates \cite{11086606}. Each variable was normalized using Min-Max normalization according to the training set statistics. Then the top model is selected to see the effect of adding the physics constraints on the forecasting performance at different sequence and prediction lengths.  

\subsection{Forecasting Results}
Table \ref{tab:avg_metrics_seq240_360_480_600} shows a clear separation between lightweight linear MLP baselines (DLinear/PI-DLinear) and the majority of transformer variants across all four history/sequence lengths.

\begin{table*}[!htbp]
\centering
\tiny
\caption{Average performance across prediction horizons $L\in\{5,10,20,40,80\}$\,\unit{\minute} for \textbf{multivariate} forecasting on MIT Supercloud under different history lengths ($T \in \{240,360,480,600\}$\,\unit{\minute}) for the proposed and baseline models. It is apparent that the proposed PI-DLinear model not only outperforms the baseline models in terms of performance, but it (the error values) is relatively stable as the sequence length increases (see Fig. \ref{fig:heatmapsMAPEMSE}) as well, a feature that is crucial when deploying such a model in data center control units, where multi-scaling and model flexibility is required at any given history and time step. The models were tested with 5 seeds, yet the standard deviation across runs for all models was negligible.}
\label{tab:avg_metrics_seq240_360_480_600}
\setlength{\tabcolsep}{3pt}
\renewcommand{\arraystretch}{1}
\begin{tabular}{l
                S[table-format=1.4] S[table-format=1.4] S[table-format=1.4] S[table-format=1.4]
                S[table-format=1.4] S[table-format=1.4] S[table-format=1.4] S[table-format=1.4]
                S[table-format=1.4] S[table-format=1.4] S[table-format=1.4] S[table-format=1.4]
                S[table-format=1.4] S[table-format=1.4] S[table-format=1.4] S[table-format=1.4]}
\toprule
& \multicolumn{4}{c}{$T=240$\,\unit{\minute}} & \multicolumn{4}{c}{$T=360$\,\unit{\minute}} & \multicolumn{4}{c}{$T=480$\,\unit{\minute}} & \multicolumn{4}{c}{$T=600$\,\unit{\minute}} \\
\cmidrule(lr){2-5}\cmidrule(lr){6-9}\cmidrule(lr){10-13}\cmidrule(lr){14-17}
\textbf{Model}
& {\textbf{MAE}} & {\textbf{MSE}} & {\textbf{MAPE}} & {\textbf{RMSE}}
& {\textbf{MAE}} & {\textbf{MSE}} & {\textbf{MAPE}} & {\textbf{RMSE}}
& {\textbf{MAE}} & {\textbf{MSE}} & {\textbf{MAPE}} & {\textbf{RMSE}}
& {\textbf{MAE}} & {\textbf{MSE}} & {\textbf{MAPE}} & {\textbf{RMSE}} \\
\midrule
iTransformer                & 0.1481 & 0.1636 & 1.1323 & 0.4000 & 0.1527 & 0.1665 & 1.1603 & 0.4039 & 0.1533 & 0.1694 & 1.1537 & 0.4071 & 0.1555 & 0.1708 & 1.2052 & 0.4087 \\
Transformer                 & 0.1670 & 0.1722 & 1.2211 & 0.4095 & 0.1726 & 0.1674 & 1.4009 & 0.4058 & 0.1754 & 0.1727 & 1.3291 & 0.4115 & 0.1785 & 0.1747 & 1.3318 & 0.4140 \\
TimeXer                      & 0.1474 & 0.1580 & 1.1021 & 0.3941 & 0.1631 & 0.1664 & 1.2364 & 0.4053 & 0.1511 & 0.1637 & 1.1139 & 0.4014 & 0.1636 & 0.1674 & 1.2195 & 0.4065 \\
TiDE                         & \underline{0.1422} & 0.1561 & 1.0628 & 0.3912 & 0.1436 & 0.1578 & 1.0743 & 0.3933 & 0.1441 & 0.1585 & 1.0708 & 0.3943 & 0.1451 & 0.1584 & 1.0860 & 0.3941 \\
TSMixer                      & 0.1507 & 0.1605 & 1.0999 & 0.3968 & 0.1547 & 0.1654 & 1.1798 & 0.4030 & 0.1549 & 0.1657 & 1.1491 & 0.4036 & 0.1640 & 0.1697 & 1.2085 & 0.4082 \\
Reformer                     & 0.1739 & 0.1700 & 1.2483 & 0.4077 & 0.1730 & 0.1666 & 1.3543 & 0.4054 & 0.1693 & 0.1759 & 1.2455 & 0.4166 & 0.1860 & 0.1775 & 1.5094 & 0.4158 \\
PatchTST                     & 0.1495 & 0.1628 & 1.1203 & 0.3993 & 0.1520 & 0.1599 & 1.1443 & 0.3959 & 0.1522 & 0.1631 & 1.1335 & 0.3999 & 0.1558 & 0.1645 & 1.1635 & 0.4016 \\
\makecell{Nonstationary\\Transformer}   & 0.1489 & 0.1736 & 1.1044 & 0.4136 & 0.1505 & 0.1752 & 1.1388 & 0.4159 & 0.1527 & 0.1674 & 1.1212 & 0.4061 & 0.1580 & 0.1702 & 1.2052 & 0.4100 \\
LightTS                      & 0.1496 & 0.1668 & 1.1490 & 0.4052 & 0.1526 & 0.1639 & 1.1687 & 0.4010 & 0.1539 & 0.1669 & 1.1695 & 0.4052 & 0.1563 & 0.1691 & 1.1749 & 0.4079 \\
FiLM                         & 0.1432 & 0.1571 & 1.0708 & 0.3925 & 0.1437 & 0.1577 & 1.0810 & 0.3931 & \underline{0.1440} & 0.1578 & 1.0840 & 0.3934 & \underline{0.1440} & 0.1582 & 1.0877 & 0.3938 \\
FEDformer                    & 0.2164 & 0.2013 & 1.7254 & 0.4478 & 0.2747 & 0.2549 & 2.3160 & 0.5044 & 0.3028 & 0.2976 & 2.5071 & 0.5455 & 0.2741 & 0.2669 & 2.2863 & 0.5151 \\
Pyraformer                   & 0.1498 & 0.1610 & 1.2292 & 0.3986 & 0.1531 & 0.1594 & 1.2034 & 0.3959 & 0.1555 & 0.1647 & 1.1664 & 0.4032 & 0.1710 & 0.1822 & 1.5822 & 0.4179 \\
DLinear                      & \textbf{0.1420} & \underline{0.1556} & \underline{1.0403} & \underline{0.3907} & \textbf{0.1426} & \underline{0.1561} & \underline{1.0484} & \underline{0.3913} & \textbf{0.1434} & \underline{0.1571} & \underline{1.0516} & \underline{0.3925} & \textbf{0.1439} & \underline{0.1576} & \underline{1.0646} & \underline{0.3931} \\
Crossformer                  & 0.1491 & 0.1564 & 1.1437 & 0.3925 & 0.1833 & 0.2650 & 2.3161 & 0.5084 & 0.1504 & 0.1579 & 1.1648 & 0.3948 & 0.1535 & 0.1628 & 1.1889 & 0.3995 \\
NLinear                      & 0.1673 & 0.2080 & 2.0779 & 0.4505 & 0.1460 & 0.1592 & 1.1018 & 0.3952 & 0.1472 & 0.1606 & 1.1004 & 0.3969 & 0.1480 & 0.1602 & 1.1144 & 0.3965 \\
Linear                       & 0.1463 & 0.1600 & 1.0771 & 0.3965 & 0.1468 & 0.1603 & 1.0863 & 0.3969 & 0.1479 & 0.1614 & 1.0834 & 0.3983 & 0.1482 & 0.1611 & 1.0906 & 0.3979 \\
\textbf{\makecell{PI-DLinear\\(Ours)}}   & \textbf{0.1420} & \textbf{0.1546} & \textbf{1.0315} & \textbf{0.3895} & \underline{0.1429} & \textbf{0.1549} & \textbf{1.0358} & \textbf{0.3899} & \textbf{0.1434} & \textbf{0.1559} & \textbf{1.0414} & \textbf{0.3910} & \textbf{0.1439} & \textbf{0.1561} & \textbf{1.0510} & \textbf{0.3914} \\
\bottomrule
\end{tabular}
\footnotesize{\textit{Note:} \textbf{Bold} indicates the best (lowest) value and \underline{underlined} indicates the second-best (lowest) value for each metric (normalized).}
\vspace{-\baselineskip}
\end{table*}

\begin{figure}[!b]
    \centering
    \begin{subfigure}[t]{0.495\linewidth}
        \centering
        \includegraphics[width=\linewidth]{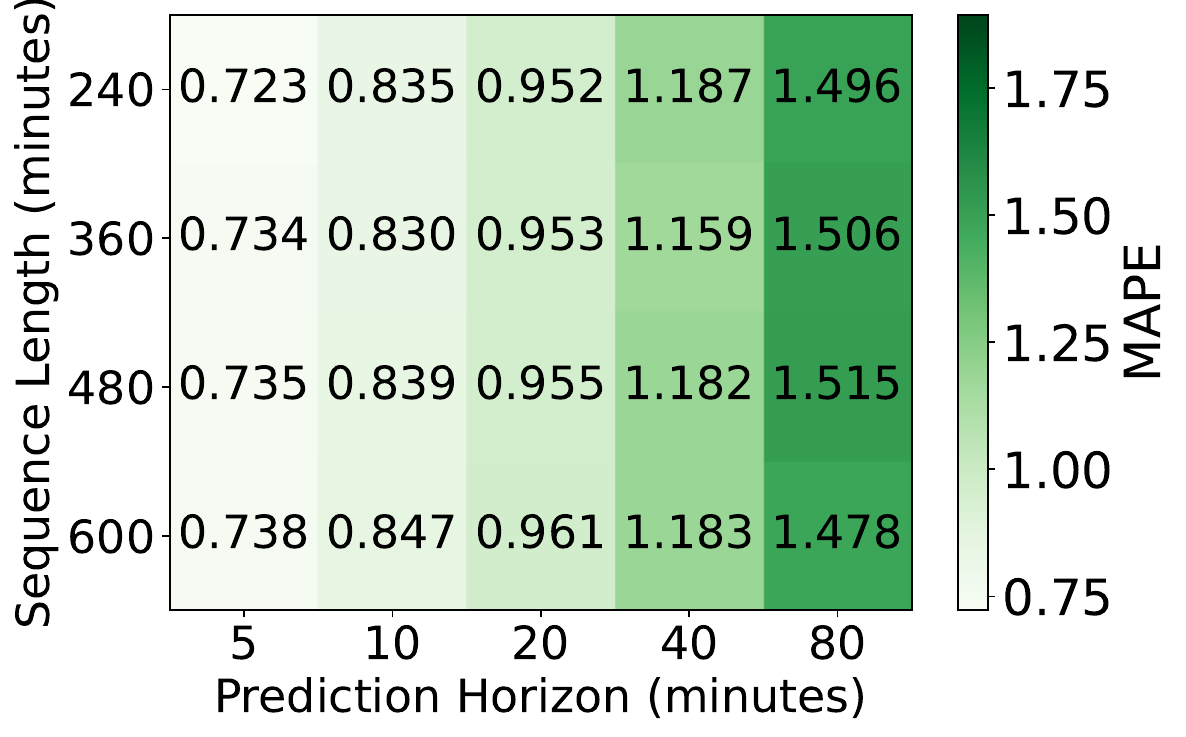}
        \caption{MAPE values.}
        \label{fig:sub1}
    \end{subfigure}
    \hfill
    \begin{subfigure}[t]{0.495\linewidth}
        \centering
        \includegraphics[width=\linewidth]{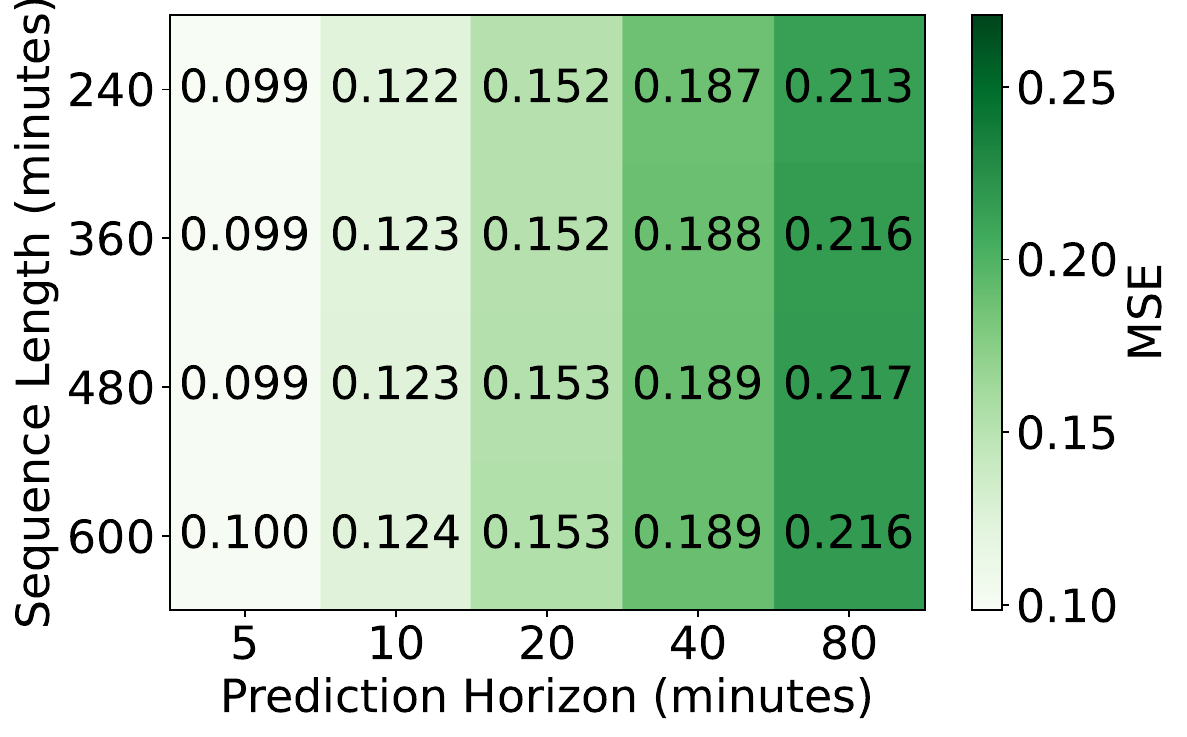}
        \caption{MSE values.}
        \label{fig:sub3}
    \end{subfigure}

    \caption{PI-DLinear forecasting results shown as heatmaps for input sequence length $T\in\{240,360,480,600\}\,\unit{\minute}$ and prediction length $L\in\{5,10,20,40,80\}\,\unit{\minute}$. PI-DLinear prediction error is stable for different sequence lengths and a shorter prediction horizon is more accurate than longer prediction horizons.}
    \label{fig:heatmapsMAPEMSE}
    \vspace{-\baselineskip}
\end{figure}
Firstly, the performance is remarkably stable as the history window grows from 240 to 600 minutes for the strongest methods. DLinear varies only slightly (MAE: $0.1420\rightarrow0.1439$; MSE: $0.1556\rightarrow0.1576$), indicating that the signal required for short-to-mid horizon forecasting is largely captured within a few hours of context. The same stability holds for TiDE and FiLM (e.g., FiLM MAE stays $\approx 0.143-0.144$ with MSE $\approx 0.157-0.158$), suggesting that extending history for this dataset diminishes the performance once the recent operating regime is observed.

Second, PI-DLinear (ours) is consistently the strongest across the error metrics and history lengths, with the most consistent gains appearing in MSE/RMSE, which is typically the most sensitive to costly spikes as shown in Table \ref{tab:avg_metrics_seq240_360_480_600}. In fact, PI-DLinear achieves the best MSE at every sequence length (0.1546, 0.1549, 0.1559, 0.1561 for 240/360/480/600 min), and correspondingly the best RMSE ($\approx 0.3895-0.3914$), which is further substantiated by the forecasting results observed in Fig. \ref{fig:teaser}. For MAE, PI-DLinear remains tied with DLinear at 240 min (0.1420) and stays within a very small margin of the best thereafter (e.g., 0.1430 at 360 min, 0.1434 at 480 min), while still reducing squared-error measures. This behavior is consistent with the physics term acting as a regularizer that suppresses high-variance deviations without over-smoothing the median error. 

In light of this, PI-DLinear encodes a mechanistic state-space model for the data center’s thermal subsystem, where GPU and memory temperatures are treated as physical state variables whose evolution must satisfy energy conservation plus Newtonian cooling in a coupled lumped RC network (Equation (\ref{eq:dpdt_constraint1})). This turns forecasting into a constrained dynamical inference problem, where the predicted power trajectories are only acceptable insofar as they adhere to a physically realizable temperature response governed by the thermal capacitances/resistances, rather than arbitrary temporal patterns. To this end, because the trace observes total power $P(t)$ but not its GPU/memory split, introducing the latent partition $\alpha$ (Equation (\ref{eq:power_split})) makes the model implicitly perform system identification of how workload power is allocated across the thermal components, guiding the learned forecast to an interpretable physical decomposition rather than purely statistical smoothing. 

Moreover, in Fig. \ref{fig:heatmapsMAPEMSE}, it is apparent that the proposed PI-DLinear prediction error seems to be independent of the sequence length and a shorter prediction horizon is more accurate than longer prediction horizons, which is expected as the error accumulates as we predict further in time. Nevertheless, such stability (as the sequence length increases) is a crucial feature of the proposed model, especially for deploying it in data center control units, where multi-scaling and model flexibility are required at any given history and time step.

A third takeaway concerns model efficiency vs. accuracy trade-offs, where several complex architectures do not translate their capacity into better average forecasting error on this workload. For example, iTransformer/TimeXer/PatchTST yield MAE values ranging $\approx 0.149-0.156$ and MSE $\approx 0.160-0.171$ depending on the sequence length, consistently behind the proposed PI-DLinear and baseline DLinear model. Meanwhile, FEDformer is a clear outlier with substantially larger errors, especially at longer histories (e.g., MAE 0.2747–0.3028 and MAPE $>2.3$ for 360–600 min), suggesting mismatch during short-term, high-variability dynamics of GPU traces. Furthermore, Crossformer exhibits a similar instability at 360 min (MSE 0.2650, MAPE 2.3161), even though it recovers at 480–600 min, highlighting that some architectures can be sensitive to the context/sequence length horizons selected.

Overall, the results support a simple conclusion aligned with the paper’s message: physics-aware linear forecasting offers a strong robustness prior for AI data center GPU power/thermal proxies. PI-DLinear performs the best relative to the strongest SOTA baselines, while systematically improving the forecasting performance across all history/sequence lengths, which is precisely the regime that matters when avoiding large prediction misses under abrupt AI workload shifts and periodic maintenance activities. The motivation for physics-aware forecasting is also brought about in this case to show that GPU power traces exhibit constrained, regime-dependent behavior driven by device management. In particular, many compute-intensive V100 jobs operated near a tight maximum-power band around the nominal power class ($\approx250$\,\unit{\watt}), consistent with power-limiting mechanisms documented by NVIDIA (i.e., performance/power capping). Such caps introduce a saturation regime where power no longer scales linearly with workload, creating non-stationary dynamics that can degrade purely data-driven linear models. This motivates incorporating physically informed constraints (as in PI-DLinear) to improve robustness and generalization under operational limits.

Furthermore, the combination of dense sampling and extended temporal coverage implies substantial non-stationarity (e.g., workload shifts, diurnal/weekly cycles, and IT maintenance activities), making the benchmark more representative of real data-center conditions than short or sparsely sampled traces. This scale also stresses model efficiency, since training and inference must operate over hundreds of thousands of time points while maintaining accuracy across different prediction horizons, which further substantiates the reason why Physics-informed DLinear (PI-DLinear) was selected, as it offers high performance and low computational complexity, making it deployable to the control and monitoring units and/or microcontrollers available at AI data centers.

\subsection{Computational Burden}
Table \ref{tab:model_cost} highlights a clear efficiency–capacity trade-off. FiLM is the most expensive option, with 12.9M parameters, 271.38 s per epoch, and 49.30 MB memory, approximately 134× more parameters, 26× longer runtime, and 131× higher memory than DLinear. TiDE sits in the middle (1.62M params, 29.10 s, 6.21 MB), remaining substantially heavier than DLinear (about 17× more parameters and 2.8× slower). In contrast, DLinear is the lightest (96k params, 10.43 s, 0.376 MB) and more importantly, PI-DLinear preserves the same parameter count and memory footprint as DLinear (96k, 0.376 MB) while increasing runtime to 20.27 s (about 1.9× slower), indicating that the physics-aware component adds compute overhead without increasing model size because it's only used in optimizing training and not during inference.
\begin{figure}[!b]
\vspace{-\baselineskip}
    \centering
    \includegraphics[width=0.75\linewidth]{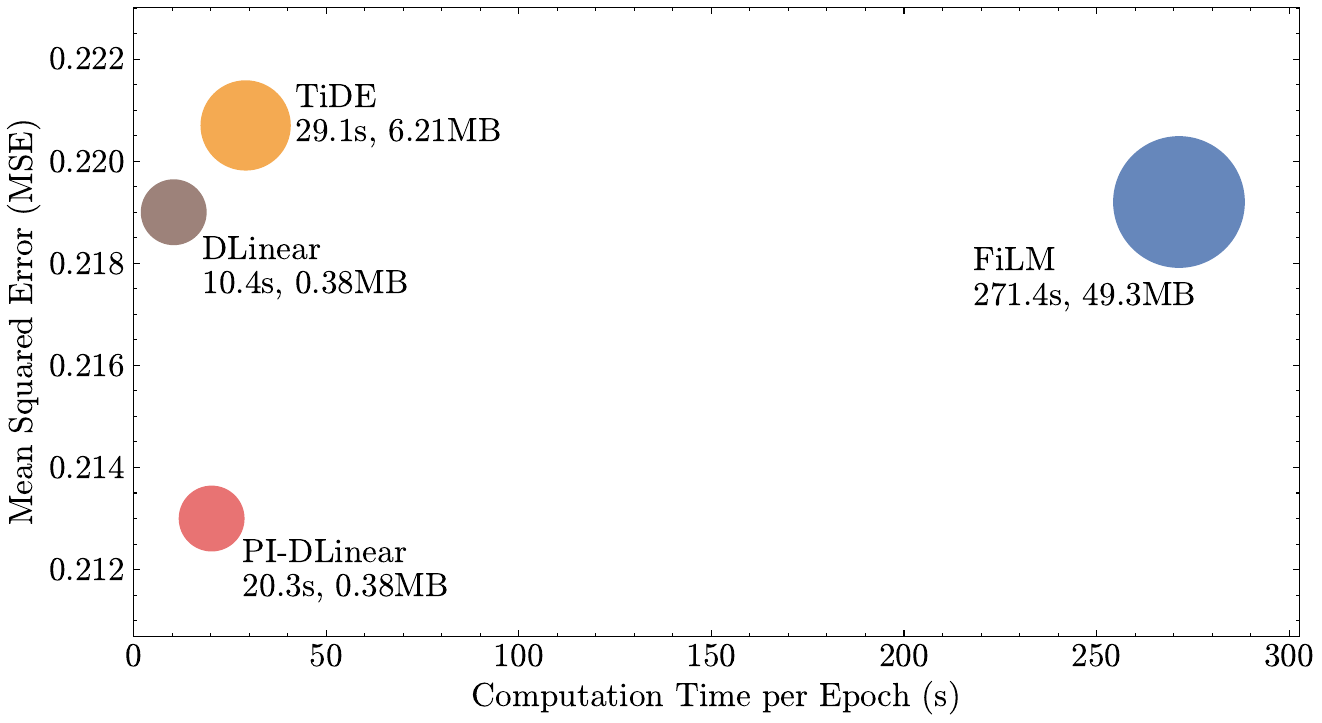}
    \caption{Bubble chart showing model efficiency comparison under input-240-predict-80 for the top 4 models. PI-Dlinear has the lowest MSE error with a computational time slightly higher than baseline DLinear. GPU model used: NVIDIA GeForce RTX 4080, CUDA version: 12.9. The model size was obtained from the checkpoint MB saved on disk.}
    \label{fig:bubble}
\end{figure}
The bubble chart in Fig. \ref{fig:bubble} indicates that PI-DLinear offers the best accuracy–efficiency trade-off among these four top models. It achieves the lowest MSE (0.213) at 20.27 s/epoch with a very small memory footprint (0.376\,\unit{\mega\byte}), suggesting that incorporating the physics-aware term improves fit without increasing model size. Relative to vanilla DLinear (MSE 0.2190, 10.43 s/epoch, 0.376\,\unit{\mega\byte}), PI-DLinear reduces the error by 0.006 MSE ($\approx$ 3\%) for this case (input-240-predict-80), at the cost of 1.94× higher compute time, consistent with added physics-related computation.
\begin{table}[!t]
\centering
\small
\caption{Model complexity and inference efficiency for a single forward pass, reported in terms of parameter count, runtime (seconds), and peak memory usage (MB).}
\label{tab:model_cost}
\setlength{\tabcolsep}{8pt}
\renewcommand{\arraystretch}{1.15}

\begin{tabular}{l 
                S[table-format=8] 
                S[table-format=3.2] 
                S[table-format=2.3]}
\toprule
\textbf{Model} 
& {\textbf{\#Params}} 
& {\textbf{Time (\unit{\second})}} 
& {\textbf{Memory (\unit{\mega\byte})}} \\
\midrule

TiDE        & 1624359   & 29.10  & 6.210 \\
FiLM        & 12923662  & 271.38 & 49.30 \\
DLinear     & 96160     & 10.43  & 0.376 \\
PI-DLinear  & 96160     & 20.27  & 0.376 \\

\bottomrule
\end{tabular}
\vspace{-1\baselineskip}
\end{table}
In contrast, TiDE is both slower (29.10 s/epoch) and less accurate (0.2207 MSE) than PI-DLinear, and requires substantially more memory (6.21\,\unit{\mega\byte}), which aligns with its heavier encoder–decoder MLP design compared to linear baselines. Finally, FiLM is an extreme outlier in cost, 271.38 s/epoch and 49.30\,\unit{\mega\byte}, yet delivers no accuracy benefit (MSE 0.2192, essentially comparable to DLinear), making it dominated in this setting, where it consumes orders of magnitude more compute/memory without improving (decreasing) the forecasting error.

\subsection{Predicting Power Throttling}
\subsubsection{Throttle Event Detection} Table~\ref{tab:throttle_results} in the appendix (Section \ref{sec:powerthrottle}) presents the throttle detection performance across different sequence lengths and prediction horizons. PI-DLinear consistently outperforms the baseline DLinear in detecting power throttling events, achieving improved detection rates in the configurations tested. The most substantial improvement occurs at sequence length $L=360$ with prediction horizon $T=10$, where PI-DLinear achieves a detection rate of 85.03\% compared to 65.29\% for DLinear, an improvement of 19.75\%. However, at very short horizons ($T=5$), DLinear actually outperforms PI-DLinear in detection rate. This suggests the physics constraints are most beneficial for longer prediction horizons ($T\geq10$), where the thermal dynamics have more time to manifest. This is further substantiated by the interesting observation, which is that PI-DLinear reaches near-perfect detection at longer horizons, achieving 99.12\% detection rate at $L=480$, $H=80$ compared to 96.27\% for DLinear. Overall, on average across all configurations, PI-DLinear improves the throttle detection rate by 6.88\%, demonstrating that the physics-informed constraints effectively encode the relationship between sustained high utilization and impending power reduction.

\subsubsection{Throttle Event Prediction Accuracy} Beyond detection rate, we evaluate prediction accuracy specifically at throttle event timesteps using Throttle MAE and Throttle RMSE. PI-DLinear achieves an average Throttle RMSE improvement of 3.92\% across all prediction lengths, with the most significant improvement of 28.61\% observed at $L=360$, $T=20$ (RMSE reduced from 0.396 to 0.283 in normalized units, corresponding to a reduction from 1,256~W to 897~W in original units). This improvement further underscores that physics-informed thermal constraints enable the model to better anticipate the magnitude of power drops during throttling events. While Throttle MAE shows marginal average improvement (0.15\%), this is expected as MAE is less sensitive to large prediction errors that occur during sudden power transitions. Yet, the combination of higher detection rates and lower RMSE indicates that PI-DLinear not only identifies more throttling events but also predicts their severity more accurately, validating the effectiveness of incorporating thermal physics and utilization-based constraints into the forecasting framework.

\subsection{Ablation Study}
Across all horizons, adding physics-aware constraints consistently improves forecasting performance relative to the state-of-the-art DLinear, with the gains becoming more noticeable as the prediction window grows, further demonstrating the potential of the proposed model for long-term power forecasting. Even with a fixed physics weight ($\lambda=0.005$), PI-DLinear reduces both MAPE and MSE at every horizon (e.g., MAPE drops from 1.5250 to 1.5088 at prediction length of 80\,\unit{\minute}), indicating that the newly derived second-order RC constraint acts as a useful regularizer for power-thermal dynamics. Ultimately, the self-adaptive weighting variant delivers the best and most consistent improvements, achieving the best scores for every prediction length averaged over the investigated look-back windows and providing the largest long-horizon performance enhancement (e.g., MAPE 1.4987 and MSE 0.2155 at a prediction length of 80\,\unit{\minute}) as outlined in Table \ref{tab:ablation_horizons}, which suggests that dynamically balancing data fit and physics residuals is important when error accumulation and regime shifts (i.e., throttling-related transients) become more prevalent at longer horizons. This is further corroborated by the large improvements observed in Fig. \ref{fig:teaser}, where the physics and throttle constraints encoded into the loss function of DLinear, delivered (1) improved transient recovery forecasting performance under abrupt AI load fluctuations, (2) stable predictions even after power throttling subsides, and (3) allowed for capturing GPU power throttling events. 

Lastly, considering the derived ODE solution alone, which considers power and thermal behavior, inertia (time constant) and coupling effects are therefore inherently present in the data center. Therefore, over longer prediction horizons, these slower physical mechanisms become more prominent, so the proposed ODE-based reconstruction was able to produce better global trajectory forecasts, which led to an improvement in the MSE error and became more competitive with the other models in Table \ref{tab:ablation_horizons}. Such results further substantiate the viability of the derived ODEs, qualifying them to be integrated as the Physics component within the proposed PI-DLinear model. 

\begin{table*}[!t]
\centering
\caption{Ablation study across the investigated forecasting horizons for multivariate power forecasting of AI data centers. We report MAPE and MSE (lower is better) for DLinear, PI-DLinear (fixed $\lambda$), and self-adaptive PI-DLinear.}
\label{tab:ablation_horizons}
\setlength{\tabcolsep}{3.5pt}
\renewcommand{\arraystretch}{1.15}

\begin{tabular}{l
                S[table-format=1.4] S[table-format=1.4]
                S[table-format=1.4] S[table-format=1.4]
                S[table-format=1.4] S[table-format=1.4]
                S[table-format=1.4] S[table-format=1.4]
                S[table-format=1.4] S[table-format=1.4]}
\toprule
& \multicolumn{2}{c}{\textbf{Horizon 5}} 
& \multicolumn{2}{c}{\textbf{Horizon 10}} 
& \multicolumn{2}{c}{\textbf{Horizon 20}} 
& \multicolumn{2}{c}{\textbf{Horizon 40}} 
& \multicolumn{2}{c}{\textbf{Horizon 80}} \\
\cmidrule(lr){2-3}\cmidrule(lr){4-5}\cmidrule(lr){6-7}\cmidrule(lr){8-9}\cmidrule(lr){10-11}
\textbf{Model} 
& {\textbf{MAPE}} & {\textbf{MSE}}
& {\textbf{MAPE}} & {\textbf{MSE}}
& {\textbf{MAPE}} & {\textbf{MSE}}
& {\textbf{MAPE}} & {\textbf{MSE}}
& {\textbf{MAPE}} & {\textbf{MSE}} \\
\midrule
Derived ODE Solution
& 0.7409 & 0.1002
&  0.8420& 0.1233
& 0.9574 & 0.1525
&1.1963& \underline{0.1886}
& 1.5252 &\textbf{0.2150}\\
DLinear
& 0.7331 & 0.0994
&  0.8422& 0.1236
& 0.9606 & 0.1531
&1.1951 & 0.1897
& 1.5250 &0.2172 \\
PI-DLinear + Constant $\lambda$
&\underline{0.7317}  & \underline{0.0993}
& \underline{0.8407} &\underline{0.1234}
&\underline{0.9561} & \underline{0.1527}
& \underline{1.1939} & 0.1893
& \underline{1.5088} & \underline{0.2165} \\
Self-Adaptive PI-DLinear
& \textbf{0.7309} & \textbf{0.0992}
& \textbf{0.8377} & \textbf{0.1232}
& \textbf{0.9554} & \textbf{0.1526}
& \textbf{1.1780} & \textbf{0.1886}
& \textbf{1.4987} &\underline{0.2155} \\
\bottomrule
\end{tabular}

\vspace{0.35em}
\footnotesize{\textbf{Note:} Bold denotes the best result per column; \underline{underline} denotes the second best. The values are averaged over each look-back window.}
\end{table*}

\section{Conclusion}
\label{conc_sec}
In this work, we addressed short-term AI data-center power forecasting under highly transient GPU workloads by introducing PI-DLinear, a physics-aware variant of DLinear that embeds a multi-node lumped thermal RC network consistent with Newton's law of cooling as a differentiable loss constraint. The model links the power drawn to GPU/memory thermal dynamics, thereby improving generalization across various history/lookback windows and forecasting horizons. Moving on, regarding the MIT Supercloud trace, PI-DLinear consistently outperformed state-of-the-art transformer and non-transformer baselines across multiple look-back lengths and horizons, with the most reliable gains appearing in MSE/MAPE/RMSE, the metrics most sensitive to the inherent costly spikes in the data. More importantly, PI-DLinear preserves DLinear's lightweight footprint, incurring overhead primarily during training due to the added physics term, which supports practical deployment in monitoring/control units present in AI data centers, where efficiency matters. Finally, the ablation study highlights that physics weighting is essential, since overly strong regularization can degrade forecasting accuracy by over-constraining the data-driven optimum, especially when considering power throttling and load transient events.

\bibliographystyle{unsrtnat}
\bibliography{references} 

\appendix
\section{Physics-Informed Neural Networks}
During forward modelling, the initial and boundary conditions are assumed to be known, bringing us to the common ODE solving problem, where PINN is leveraged as a conventional numerical solver approach. With that, let the unknown $u(\mathbf{x},t)$ be determined by an \textit{ansatz} $u(\mathbf{x},t; \mathbf{w})$, comprising a NN with input features $\mathbf{x}= (x_1,...,x_d)$, time $t$ to study the temporal correlations, and network weights $\mathbf{w}$. Merely  one output is approximated (regression problem) as $u$ is a scalar function. However, given a ODE/PDE system, multiple outputs can be acquired, one for each variable. To this end, the objective is to obtain network weights $\mathbf{w}$ such that \cite{MCCLENNY2023111722}
\begin{equation}
\begin{split}
&F(u(\mathbf{x},t;\mathbf{w}),\mathbf{x},t,Du(\mathbf{x},t;\mathbf{w}),D^2u(\mathbf{x},t;\mathbf{w}),...,\\
&D^ku(\mathbf{x},t;\mathbf{w});\lambda)=f(\mathbf{x},t).\hspace{0.15in} \mathbf{x},t\in \Omega
\label{eq:Def1}
\end{split}
\end{equation}
The partial derivatives $D^\alpha u(\mathbf{x}; \mathbf{w})$ of the NN output, for $\mathbf{x}$ and $\mathbf{w}$ at time $t$ in function $F$, is calculated using automatic differentiation methods. Therefore, it is worth noting the loss functions as \cite{MCCLENNY2023111722}
\begin{equation}
\label{eq:Def22}
\begin{split}
\mathcal{L}_u(\mathbf{w})
&= \int_{\Omega} \left| \hat{u}(\mathbf{x},t) - u(\mathbf{x},t) \right|^2 \, d\mathbf{x}, \\
\mathcal{L}_r(\mathbf{w},\lambda)
&= \int_{\Omega} \Bigl| 
F\bigl(
u(\mathbf{x},t;\mathbf{w}), \mathbf{x}, t, Du(\mathbf{x},t;\mathbf{w}), \\
&\qquad D^2u(\mathbf{x},t;\mathbf{w}), \ldots, D^k u(\mathbf{x},t;\mathbf{w}); \lambda
\bigr)
- f(\mathbf{x},t)
\Bigr|^2 \, d\mathbf{x}.
\end{split}
\end{equation}
$\mathcal{L}_u$ is the mean-squared error ($\mathrm{MSE}_u$) that the NN incorporates to forecast the initial and boundary conditions, as well as utilizing training data for calibration represented via $\left\{\mathbf{x}_u^i,t_u^i,u^i\right\}_{i=1}^{N_u}$, with $\hat{u}$ defined as the predicted label based on the input $\mathbf{x}_u^i$ at time $t_u^i$ and compared with the true label $u^i$. $\mathcal{L}_r$ is the $\mathrm{MSE}_r$, which consists of evaluations of the residual/physics function $f$ over $\left\{\mathbf{x}_r^i,t_r^i\right\}_{i=1}^{N_r}$ with input $\mathbf{x}_r^i$ at time $t_f^i$ \cite{Deng2023physics}.
Conventionally, the integrals in Equation (\ref{eq:Def22}) is numerically approximated by constructing Monte-Carlo estimates through uniformly distributed random samples $\{\mathbf{x}_r^i,t_r^i\}_{i=1}^{N_r}\subset\Omega$ and $\{\mathbf{x}_u^i,t_u^i\}_{i=1}^{N_u}\subset\Omega$ as \cite{MCCLENNY2023111722}
\vspace*{-0.2\baselineskip}
\begin{equation}
\label{eq:Def3}
\begin{split}
\mathcal{L}_u(\mathbf{w})
&\doteq \frac{1}{N_u}\sum_{i=1}^{N_u}
\left|\hat{u}(\mathbf{x}_u^i,t_u^i)-u(\mathbf{x}_u^i,t_u^i)\right|, \\
\mathcal{L}_r(\mathbf{w})
&\doteq \frac{1}{N_r}\sum_{i=1}^{N_r}
\Bigl|
F\bigl(
u(\mathbf{x}_r^i,t_r^i;\mathbf{w}), \mathbf{x}_r^i, t_r^i,
Du(\mathbf{x}_r^i,t_r^i;\mathbf{w}), \\
&\qquad D^2u(\mathbf{x}_r^i,t_r^i;\mathbf{w}), \ldots,
D^m u(\mathbf{x}_r^i,t_r^i;\mathbf{w}); \lambda
\bigr)
- f(\mathbf{x}_r^i,t_r^i)
\Bigr|^2.
\end{split}
\end{equation}

\section{Methodology}
\subsection{Problem Statement}
We address the problem of short-term power forecasting for GPU-accelerated data centers. Given a multivariate time series of GPU telemetry
\begin{equation}
    \mathbf{X} = \{\mathbf{x}_1, \mathbf{x}_2, \ldots, \mathbf{x}_L\} \in \mathbb{R}^{L \times C},
\end{equation}
where $L$ is the historical lookback window and each observation $\mathbf{x}_t \in \mathbb{R}^D$ comprises $C=5$ features/co-variates
\begin{equation}
\mathbf{x}_t = \begin{bmatrix} u_t^{(g)} & u_t^{(m)} & T_t^{(g)} & T_t^{(m)} & P_t \end{bmatrix}^\top,
\end{equation}
representing GPU utilization ($u^{(g)}$), memory utilization ($u^{(m)}$), GPU temperature ($T^{(g)}$), memory temperature ($T^{(m)}$), and electrical power consumption ($P$), respectively.

Our objective is to predict future power consumption over a horizon $T$ as
\begin{equation}
\hat{\mathbf{y}} = \{\hat{P}_{L+1}, \hat{P}_{L+2}, \ldots, \hat{P}_{L+T}\} \in \mathbb{R}^T.
\end{equation}

\subsection{Thermal-Electrical Analogy: Theoretical Foundation}

The foundation of our physics-informed approach lies in the well-established \textit{thermal-electrical analogy}, which maps heat transfer phenomena to equivalent electrical circuits. This analogy arises from the mathematical similarity between Fourier's law of heat conduction and Ohm's law of electrical conduction.

\subsubsection{Fundamental Analogies}

The thermal-electrical analogy establishes the following correspondences.

\begin{table*}[h]
\centering
\caption{Thermal-Electrical Analogy}
\label{tab:analogy}
\begin{tabular}{lcc}
\toprule
\textbf{Quantity} & \textbf{Thermal Domain} & \textbf{Electrical Domain} \\
\midrule
Potential (Across variable) & Temperature $T$ [K] & Voltage $V$ [V] \\
Flow (Through variable) & Heat flow $\dot{Q}$ [W] & Current $I$ [A] \\
Resistance & Thermal resistance $R_{th}$ [K/W] & Electrical resistance $R$ [$\Omega$] \\
Capacitance & Thermal capacitance $C_{th}$ [J/K] & Electrical capacitance $C$ [F] \\
Source & Heat source $P$ [W] & Current source $I$ [A] \\
\bottomrule
\end{tabular}
\end{table*}

\subsubsection{Governing Equations}

\textbf{Ohm's Law (Electrical):}
\begin{equation}
V = I \cdot R \quad \Longleftrightarrow \quad I = \frac{V}{R}
\end{equation}

\textbf{Fourier's Law (Thermal):}
\begin{equation}
\Delta T = \dot{Q} \cdot R_{th} \quad \Longleftrightarrow \quad \dot{Q} = \frac{\Delta T}{R_{th}}
\end{equation}

\textbf{Capacitor Equation (Electrical):}
\begin{equation}
I = C \frac{dV}{dt}
\end{equation}

\textbf{Thermal Mass Equation:}
\begin{equation}
\dot{Q} = C_{th} \frac{dT}{dt}
\end{equation}

where the thermal capacitance $C_{th} = m \cdot c_p$ is the product of mass $m$ and specific heat capacity $c_p$.

\subsection{Lumped-Parameter Thermal Model for GPU Systems}

Modern GPUs consist of multiple thermally-coupled components. We employ a \textit{lumped-parameter} approach, which assumes spatial temperature uniformity within each component. This assumption is well-justified for GPU and memory modules given their high thermal conductivity and small dimensions.

\subsubsection{Two-Node RC Thermal Network}

We model the GPU system as a coupled two-node thermal network comprising:
\begin{itemize}
    \item \textbf{Node 1 (GPU):} Temperature $T_g$, thermal capacitance $C_g$
    \item \textbf{Node 2 (Memory):} Temperature $T_m$, thermal capacitance $C_m$
    \item \textbf{Reference (Ambient):} Temperature $T_a$ (assumed constant or slowly varying)
\end{itemize}

The thermal resistances are:
\begin{itemize}
    \item $R_{ga}$: GPU to ambient (through heatsink/airflow)
    \item $R_{ma}$: Memory to ambient
    \item $R_{gm}$: GPU to memory (thermal coupling through PCB/substrate)
\end{itemize}

\subsection{Equivalent Circuit Derivation}
\label{APP:Circuit}

\begin{figure}[!htbp]
\centering
\resizebox{3.5in}{!}{%
\begin{tikzpicture}[scale=1.0, transform shape]


\draw[thick] (0, 0) circle (0.1);
\node[above] at (0, 0.2) {$T_a$};

\draw[thick] (0.1, 0) -- (0.6, 0);
\draw[thick] (0.6, 0) -- (0.75, 0.18) -- (1.05, -0.18) -- (1.35, 0.18) -- (1.65, -0.18) -- (1.95, 0.18) -- (2.1, 0);
\draw[thick] (2.1, 0) -- (2.6, 0);
\node[above, red] at (1.35, 0.2) {$R_{ma}$};

\draw[fill=black] (2.7, 0) circle (0.08);
\node[above] at (2.7, 0.2) {$T_m$};

\draw[thick] (2.78, 0) -- (3.3, 0);
\draw[thick] (3.3, 0) -- (3.45, 0.18) -- (3.75, -0.18) -- (4.05, 0.18) -- (4.35, -0.18) -- (4.65, 0.18) -- (4.8, 0);
\draw[thick] (4.8, 0) -- (5.3, 0);
\node[above, red] at (4.05, 0.2) {$R_{gm}$};

\draw[fill=black] (5.4, 0) circle (0.08);
\node[above] at (5.4, 0.2) {$T_g$};

\draw[thick] (5.48, 0) -- (6.8, 0);

\draw[thick] (2.7, -0.08) -- (2.7, -0.5);
\draw[thick] (2.35, -0.5) -- (3.05, -0.5);
\draw[thick] (2.35, -0.65) -- (3.05, -0.65);
\draw[thick] (2.7, -0.65) -- (2.7, -0.9);
\draw[thick] (2.45, -0.9) -- (2.95, -0.9);
\draw[thick] (2.52, -0.97) -- (2.88, -0.97);
\draw[thick] (2.59, -1.04) -- (2.81, -1.04);
\node[purple] at (2.12, -0.57) {$C_m$};

\draw[thick] (5.4, -0.08) -- (5.4, -0.5);
\draw[thick] (5.05, -0.5) -- (5.75, -0.5);
\draw[thick] (5.05, -0.65) -- (5.75, -0.65);
\draw[thick] (5.4, -0.65) -- (5.4, -0.9);
\draw[thick] (5.15, -0.9) -- (5.65, -0.9);
\draw[thick] (5.22, -0.97) -- (5.58, -0.97);
\draw[thick] (5.29, -1.04) -- (5.51, -1.04);
\node[purple] at (4.83, -0.57) {$C_g$};


\draw[thick] (6.8, 0) -- (6.8, 0.5);
\draw[thick] (6.8, 0.5) -- (7.15, 0.5);
\draw[thick] (7.5, 0.5) circle (0.3);
\draw[->, thick] (7.68, 0.5) -- (7.32, 0.5);
\draw[thick] (7.8, 0.5) -- (8.1, 0.5);
\node[blue] at (8.55, 0.5) {$\alpha P$};

\draw[thick] (6.8, 0) -- (6.8, -0.5);
\draw[thick] (6.8, -0.5) -- (7.15, -0.5);
\draw[thick] (7.5, -0.5) circle (0.3);
\draw[->, thick] (7.68, -0.5) -- (7.32, -0.5);
\draw[thick] (7.8, -0.5) -- (8.1, -0.5);
\node[blue] at (8.75, -0.5) {$(1{-}\alpha) P$};

\draw[->, thick, orange] (1.0, -0.3) -- (1.7, -0.3);
\node[below, orange, font=\scriptsize] at (1.35, -0.33) {$\frac{T_m - T_a}{R_{ma}}$};

\draw[<->, thick, purple!70] (3.6, -0.3) -- (4.5, -0.3);
\node[below, purple!70, font=\scriptsize] at (4.05, -0.33) {$\frac{T_g - T_m}{R_{gm}}$};

\node[draw, rounded corners, fill=yellow!10, align=left, font=\scriptsize] at (4.0, -2.4) {
    \textbf{Thermal-Electrical Analogy:}\\
    Temperature $T$ $\leftrightarrow$ Voltage $V$\\
    Heat flow $\dot{Q}$ $\leftrightarrow$ Current $I$\\
    Thermal resistance $R_{th}$ $\leftrightarrow$ Resistance $R$\\
    Thermal capacitance $C_{th}$ $\leftrightarrow$ Capacitance $C$\\
    Heat source $P$ $\leftrightarrow$ Current source $I$
};

\node[below, font=\scriptsize, gray] at (0, -0.3) {Ambient};
\node[below, font=\scriptsize, gray] at (2.7, -1.15) {Memory};
\node[below, font=\scriptsize, gray] at (5.4, -1.15) {GPU};

\end{tikzpicture}%
}

\vspace{0.3cm}

\textbf{Governing ODEs (Kirchhoff's Current Law at each node):}
\begin{align*}
\text{GPU Node } (T_g): \quad & C_g \frac{dT_g}{dt} = \alpha P - \frac{T_g - T_a}{R_{ga}} - \frac{T_g - T_m}{R_{gm}} \\[6pt]
\text{Memory Node } (T_m): \quad & C_m \frac{dT_m}{dt} = (1-\alpha) P - \frac{T_m - T_a}{R_{ma}} + \frac{T_g - T_m}{R_{gm}}
\end{align*}

\caption{\textbf{Equivalent RC thermal circuit for coupled GPU-Memory system.} Current sources represent heat input from electrical power dissipation ($P$ split by factor $\alpha$). Capacitors represent thermal mass (ability to store thermal energy). Resistors represent thermal resistance to heat flow. Applying Kirchhoff's Current Law at each node yields the governing ODEs.}
\label{fig:rc_circuit}
\end{figure}
\subsection{Derivation of Governing ODEs via Kirchhoff's Current Law}

Applying Kirchhoff's Current Law (KCL) at each node, stating that the sum of currents entering a node equals the sum of currents leaving, we derive the governing ordinary differential equations.

\subsubsection{Node 1: GPU Die ($T_g$)}

At the GPU node, the heat balance is:
\begin{equation}
\underbrace{\dot{Q}_{\text{in}}^{(g)}}_{\text{Heat input}} = \underbrace{\dot{Q}_{\text{stored}}^{(g)}}_{\text{Heat stored}} + \underbrace{\dot{Q}_{ga}}_{\text{Heat to ambient}} + \underbrace{\dot{Q}_{gm}}_{\text{Heat to memory}}
\end{equation}

Substituting the constitutive relations:
\begin{align}
\alpha P &= C_g \frac{dT_g}{dt} + \frac{T_g - T_a}{R_{ga}} + \frac{T_g - T_m}{R_{gm}}
\end{align}

Rearranging to standard ODE form:
\begin{equation}
\boxed{C_g \frac{dT_g}{dt} = \alpha P - \frac{T_g - T_a}{R_{ga}} - \frac{T_g - T_m}{R_{gm}}}
\label{eq:ode_gpu}
\end{equation}

\subsubsection{Node 2: Memory ($T_m$)}

Similarly, at the memory node:
\begin{equation}
\underbrace{\dot{Q}_{\text{in}}^{(m)}}_{\text{Heat input}} + \underbrace{\dot{Q}_{gm}}_{\text{Heat from GPU}} = \underbrace{\dot{Q}_{\text{stored}}^{(m)}}_{\text{Heat stored}} + \underbrace{\dot{Q}_{ma}}_{\text{Heat to ambient}}
\end{equation}

Note that heat flows \textit{into} the memory node from the GPU when $T_g > T_m$:
\begin{align}
(1-\alpha) P + \frac{T_g - T_m}{R_{gm}} &= C_m \frac{dT_m}{dt} + \frac{T_m - T_a}{R_{ma}}
\end{align}

Rearranging:
\begin{equation}
\boxed{C_m \frac{dT_m}{dt} = (1-\alpha) P - \frac{T_m - T_a}{R_{ma}} + \frac{T_g - T_m}{R_{gm}}}
\label{eq:ode_mem}
\end{equation}

\begin{table*}[!htbp]
\centering
\caption{Physical interpretation of RC thermal network parameters}
\label{tab:rc_params}
\begin{tabular}{clll}
\toprule
\textbf{Symbol} & \textbf{Name} & \textbf{Units} & \textbf{Physical Meaning} \\
\midrule
$C_g$ & GPU thermal capacitance & J/K & Heat required to raise GPU temp by 1K \\
$C_m$ & Memory thermal capacitance & J/K & Heat required to raise memory temp by 1K \\
$R_{ga}$ & GPU-to-ambient resistance & K/W & Temp rise per watt dissipated to ambient \\
$R_{ma}$ & Memory-to-ambient resistance & K/W & Temp rise per watt dissipated to ambient \\
$R_{gm}$ & GPU-memory coupling resistance & K/W & Thermal coupling strength between components \\
$T_a$ & Ambient temperature & K & Reference/environment temperature \\
$\alpha$ & Power split factor & -- & Fraction of total power dissipated in GPU \\
\bottomrule
\end{tabular}
\end{table*}

\subsection{Power-Temperature Rate Relationship}

A key insight for power forecasting is that Equations~\eqref{eq:ode_gpu} and~\eqref{eq:ode_mem} can be \textit{inverted} to express power in terms of temperature dynamics. From the GPU thermal ODE, we get
\begin{equation}
C_g \frac{dT_g}{dt} = \alpha P - \frac{T_g - T_a}{R_{ga}} - \frac{T_g - T_m}{R_{gm}}.
\end{equation}

Solving for $P$:
\begin{equation}
P = \frac{1}{\alpha}\left[ C_g \frac{dT_g}{dt} + \frac{T_g - T_a}{R_{ga}} + \frac{T_g - T_m}{R_{gm}} \right]
\label{eq:power_from_temp}
\end{equation}

Differentiating both sides with respect to time yields the \textbf{power rate constraint} as
\begin{equation}
\boxed{\frac{dP}{dt} = \frac{1}{\alpha}\left[ C_g \frac{d^2T_g}{dt^2} + \frac{1}{R_{ga}}\frac{dT_g}{dt} + \frac{1}{R_{gm}}\left(\frac{dT_g}{dt} - \frac{dT_m}{dt}\right) \right]}
\label{eq:dpdt_constraint}
\end{equation}

Combining terms:
\begin{equation}
\frac{dP}{dt} = \frac{1}{\alpha}\left[ C_g \frac{d^2T_g}{dt^2} + \left(\frac{1}{R_{ga}} + \frac{1}{R_{gm}}\right)\frac{dT_g}{dt} - \frac{1}{R_{gm}}\frac{dT_m}{dt} \right]
\label{eq:dpdt_combined}
\end{equation}

\section{Power Throttling Prediction Results}
\label{sec:powerthrottle}
The power throttling predictions results are summarized in Tables \ref{tab:throttle_results} and \ref{tab:throttle_compact}.
\begin{table*}[!htbp]
\centering
\caption{\textbf{Throttle Detection Performance.} Comparison of DLinear (baseline) and PI-DLinear across different sequence lengths ($L$) and prediction horizons ($T$). Detection Rate measures the percentage of true throttle events identified. Throttle MAE and RMSE measure prediction accuracy specifically at throttle event timesteps. Bold indicates better performance. $\Delta$ columns show improvement of PI-DLinear over DLinear.}
\label{tab:throttle_results}
\resizebox{\textwidth}{!}{%
\begin{tabular}{cc|ccc|ccc|ccc}
\toprule
& & \multicolumn{3}{c|}{\textbf{Detection Rate (\%)}} & \multicolumn{3}{c|}{\textbf{Throttle MAE}} & \multicolumn{3}{c}{\textbf{Throttle RMSE}} \\
$L$ & $T$ & DLinear & PI-DLinear & $\Delta$ (pp) & DLinear & PI-DLinear & $\Delta$ (\%) & DLinear & PI-DLinear & $\Delta$ (\%) \\
\midrule
240 & 5  & \textbf{61.92} & 44.35 & -17.57 & \textbf{0.1106} & 0.1221 & -10.39 & \textbf{0.1721} & 0.2734 & -58.85 \\
240 & 10 & 65.45 & \textbf{71.52} & +6.06 & 0.1455 & \textbf{0.1398} & +3.92 & 0.3177 & \textbf{0.2925} & +7.94 \\
240 & 20 & 62.92 & \textbf{76.47} & +13.55 & 0.1638 & \textbf{0.1635} & +0.17 & 0.3790 & \textbf{0.3725} & +1.70 \\
240 & 40 & 63.85 & \textbf{74.41} & +10.56 & \textbf{0.2051} & 0.2058 & -0.36 & \textbf{0.4993} & 0.5187 & -3.89 \\
240 & 80 & 68.70 & \textbf{87.54} & +18.85 & \textbf{0.2290} & 0.2358 & -2.96 & 0.5279 & \textbf{0.5233} & +0.88 \\
\midrule
360 & 5  & \textbf{68.95} & 66.53 & -2.42 & \textbf{0.1324} & 0.1384 & -4.53 & \textbf{0.3821} & 0.3989 & -4.40 \\
360 & 10 & 65.29 & \textbf{85.03} & +19.75 & \textbf{0.1431} & 0.1565 & -9.38 & \textbf{0.3645} & 0.3973 & -9.01 \\
360 & 20 & 72.00 & \textbf{86.93} & +14.93 & 0.1627 & \textbf{0.1578} & +2.99 & 0.3962 & \textbf{0.2828} & +28.61 \\
360 & 40 & 75.00 & \textbf{85.85} & +10.85 & \textbf{0.2001} & 0.2001 & -0.01 & 0.4644 & \textbf{0.4575} & +1.49 \\
360 & 80 & 85.37 & \textbf{86.68} & +1.31 & \textbf{0.2364} & 0.2380 & -0.67 & \textbf{0.5215} & 0.5251 & -0.68 \\
\midrule
480 & 5  & \textbf{71.10} & 66.92 & -4.18 & \textbf{0.1562} & 0.1759 & -12.60 & \textbf{0.5264} & 0.6239 & -18.53 \\
480 & 10 & \textbf{81.95} & 78.40 & -3.55 & \textbf{0.1646} & 0.1727 & -4.96 & 0.4391 & \textbf{0.4355} & +0.83 \\
480 & 20 & 83.81 & \textbf{85.90} & +2.09 & 0.1782 & \textbf{0.1731} & +2.90 & 0.4322 & \textbf{0.4066} & +5.93 \\
480 & 40 & 88.26 & \textbf{95.54} & +7.28 & 0.2098 & \textbf{0.2081} & +0.77 & 0.5072 & \textbf{0.4581} & +9.67 \\
480 & 80 & 96.27 & \textbf{99.12} & +2.85 & \textbf{0.2557} & 0.2563 & -0.23 & 0.5513 & \textbf{0.5498} & +0.28 \\
\midrule
600 & 5  & \textbf{80.24} & 71.37 & -8.87 & \textbf{0.1627} & 0.1656 & -1.75 & 0.5413 & \textbf{0.4586} & +15.27 \\
600 & 10 & 83.08 & \textbf{86.40} & +3.32 & 0.1691 & \textbf{0.1670} & +1.22 & 0.4068 & \textbf{0.3825} & +5.97 \\
600 & 20 & 83.16 & \textbf{90.31} & +7.14 & \textbf{0.1721} & 0.1784 & -3.63 & 0.4105 & \textbf{0.3996} & +2.67 \\
600 & 40 & 89.29 & \textbf{91.90} & +2.62 & 0.2086 & \textbf{0.1952} & +6.41 & 0.4538 & \textbf{0.4246} & +6.43 \\
600 & 80 & 90.75 & \textbf{95.15} & +4.41 & 0.2559 & \textbf{0.2480} & +3.12 & 0.5002 & \textbf{0.4960} & +0.83 \\
\bottomrule
\end{tabular}%
}
\end{table*}

\begin{table*}[!htbp]
\centering

\caption{\textbf{Throttle Detection Performance.} Detection Rate (\%), Throttle MAE, and Throttle RMSE for DLinear (DL) and PI-DLinear (PI). Bold indicates better performance.}
\label{tab:throttle_compact}
\resizebox{\columnwidth}{!}{%
\fontsize{4}{4}\selectfont
\begin{tabular}{cc|cc|cc|cc}
\toprule
& & \multicolumn{2}{c|}{\textbf{Det. Rate (\%)}} & \multicolumn{2}{c|}{\textbf{Thr. MAE}} & \multicolumn{2}{c}{\textbf{Thr. RMSE}} \\
$L$ & $T$ & DL & PI & DL & PI & DL & PI \\
\midrule
240 & 5  & \textbf{61.9} & 44.4 & \textbf{0.111} & 0.122 & \textbf{0.172} & 0.273 \\
240 & 10 & 65.5 & \textbf{71.5} & 0.146 & \textbf{0.140} & 0.318 & \textbf{0.293} \\
240 & 20 & 62.9 & \textbf{76.5} & 0.164 & \textbf{0.164} & 0.379 & \textbf{0.373} \\
240 & 40 & 63.9 & \textbf{74.4} & \textbf{0.205} & 0.206 & \textbf{0.499} & 0.519 \\
240 & 80 & 68.7 & \textbf{87.5} & \textbf{0.229} & 0.236 & 0.528 & \textbf{0.523} \\
\midrule
360 & 5  & \textbf{69.0} & 66.5 & \textbf{0.132} & 0.138 & \textbf{0.382} & 0.399 \\
360 & 10 & 65.3 & \textbf{85.0} & \textbf{0.143} & 0.157 & \textbf{0.365} & 0.397 \\
360 & 20 & 72.0 & \textbf{86.9} & 0.163 & \textbf{0.158} & 0.396 & \textbf{0.283} \\
360 & 40 & 75.0 & \textbf{85.9} & \textbf{0.200} & 0.200 & 0.464 & \textbf{0.458} \\
360 & 80 & 85.4 & \textbf{86.7} & \textbf{0.236} & 0.238 & \textbf{0.522} & 0.525 \\
\midrule
480 & 5  & \textbf{71.1} & 66.9 & \textbf{0.156} & 0.176 & \textbf{0.526} & 0.624 \\
480 & 10 & \textbf{82.0} & 78.4 & \textbf{0.165} & 0.173 & 0.439 & \textbf{0.436} \\
480 & 20 & 83.8 & \textbf{85.9} & 0.178 & \textbf{0.173} & 0.432 & \textbf{0.407} \\
480 & 40 & 88.3 & \textbf{95.5} & 0.210 & \textbf{0.208} & 0.507 & \textbf{0.458} \\
480 & 80 & 96.3 & \textbf{99.1} & \textbf{0.256} & 0.256 & 0.551 & \textbf{0.550} \\
\midrule
600 & 5  & \textbf{80.2} & 71.4 & \textbf{0.163} & 0.166 & 0.541 & \textbf{0.459} \\
600 & 10 & 83.1 & \textbf{86.4} & 0.169 & \textbf{0.167} & 0.407 & \textbf{0.383} \\
600 & 20 & 83.2 & \textbf{90.3} & \textbf{0.172} & 0.178 & 0.411 & \textbf{0.400} \\
600 & 40 & 89.3 & \textbf{91.9} & 0.209 & \textbf{0.195} & 0.454 & \textbf{0.425} \\
600 & 80 & 90.7 & \textbf{95.2} & 0.256 & \textbf{0.248} & 0.500 & \textbf{0.496} \\
\bottomrule
\end{tabular}
}
\end{table*}

\section{Weight Visualization of PI-Dlinear}

\begin{figure*}[!htbp]
    \centering
    \begin{subfigure}{\linewidth}
        \centering
        \includegraphics[width=\linewidth]{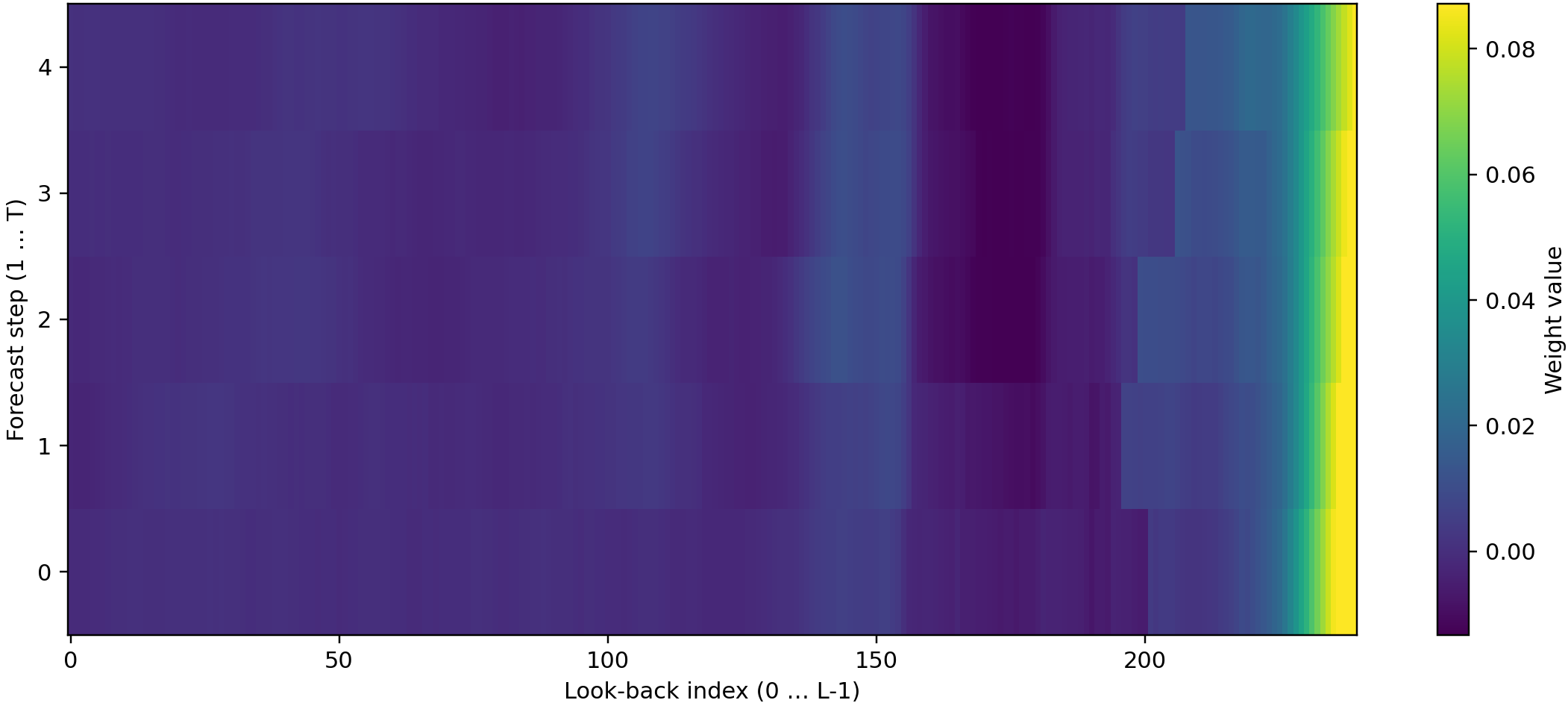}
        \label{fig:sub1}
        \caption{\textbf{Trend component}. Weight map for the trend branch, exhibiting smoother structure and stronger emphasis on recent history, consistent with a stable baseline extrapolation across horizons.}
    \end{subfigure}

    \vspace{0.1cm}

    \begin{subfigure}{\linewidth}
        \centering
        \includegraphics[width=\linewidth]{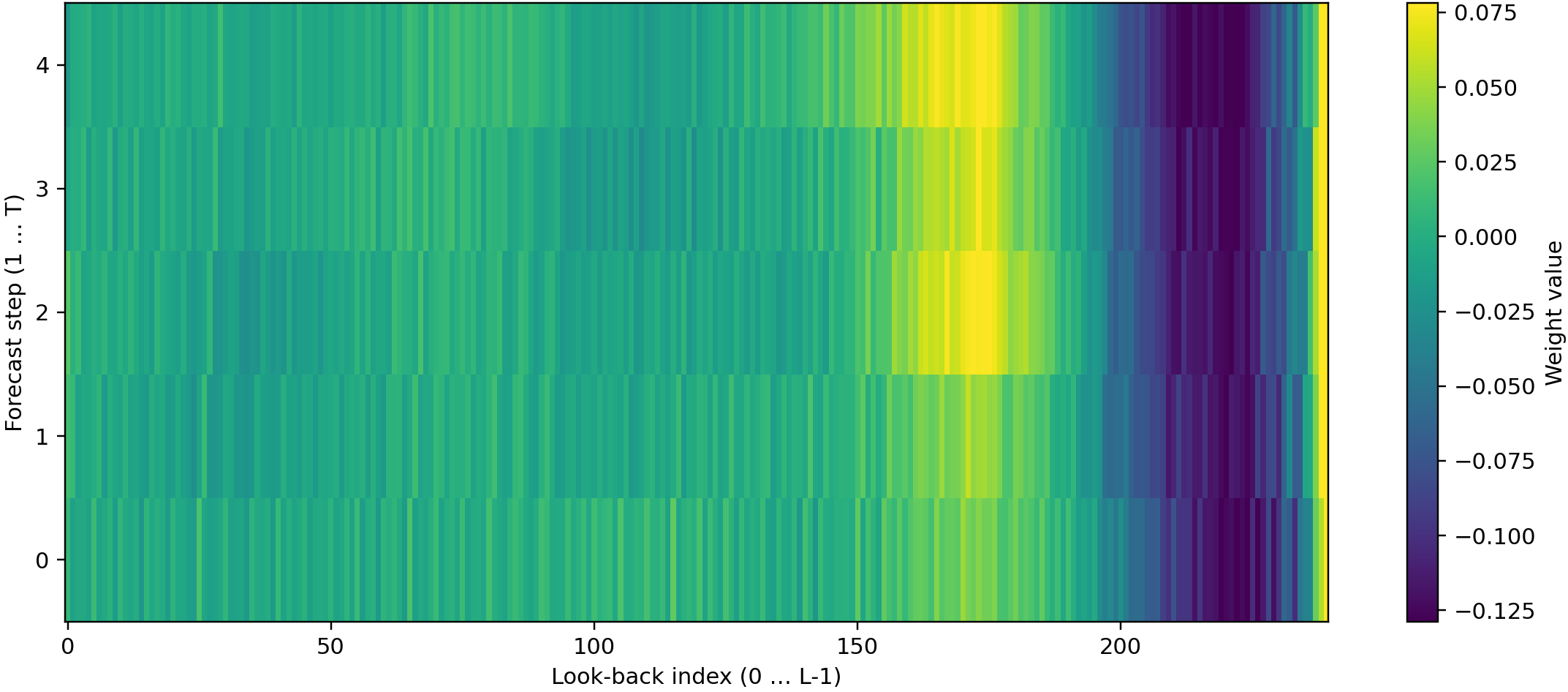}
        \label{fig:sub3}
        \caption{\textbf{Seasonal (remainder) component}. Weight map for the seasonal branch, showing structured, sign-varying lag contributions that provide horizon-dependent residual corrections.}
    \end{subfigure}

    \caption{Learned linear projection weights of PI-DLinear over the look-back window for each forecast step. The x-axis denotes the look-back index $(0,...,L-1)$ up to $L=240$\,\unit{\minute}, the y-axis denotes the forecast step $(0,...,T)$ up to $T=5$\,\unit{\minute}, and the colors encode the signed weight magnitude, indicating whether each lag contributes positively or negatively to the forecast.}
    \label{fig:heatmapsweights}
\end{figure*}

Fig. \ref{fig:heatmapsweights} presents the heatmaps of the trend and seasional components of the proposed PI-DLinear model. The heatmaps can be interpreted as time-varying linear attribution maps, where for each forecast step  $t\in \{1,...,T\}$ (y-axis), PI-DLinear assigns a weight to each look-back/sequence length index $k\in \{0,...,L-1\}$ (x-axis), and the prediction is formed by a weighted aggregation of the decomposed history. This interpretation is consistent with the DLinear family, which performs a moving-average decomposition into trend and remainder/seasonal components and then applies separate linear projections whose outputs are summed to produce the final forecast.

The seasonal heatmap shows higher spatial variability across the look-back window and more pronounced sign changes (positive vs. negative weights), consistent with a remainder branch that corrects for higher-frequency, short-term deviations rather than extrapolating a smooth trajectory. Moreover, most early-to-mid history indices contribute with relatively small magnitude (near-zero/teal color), while the model places structured emphasis on a recent sub-window, i.e. a region of positive weights (yellow-green) appears in the later part of the look-back window, followed by a strong negative band (dark purple) closer to the end of the window, and then a narrow return to positive at the extreme right edge. Functionally, this indicates a contrastive use of recent lags, boosting certain recent oscillations while suppressing others, to shape the seasonal correction, underscoring the importance of adding a physics-based regularizer/constraint to the loss function as means to improve model apprehensibility across the entire horizon of such highly variable data. Across the forecast steps, these patterns are broadly similar but not identical, suggesting the seasonal head learns horizon-dependent adjustments (i.e., the correction for step $t=1$ is not identical to that for step $t=T$), which is exactly where a seasonal/residual branch is expected to express step-specific behavior.

The trend heatmap is smoother and more coherent across the look-back axis, with substantially less fine-grained alternation than the seasonal map. The most salient feature is the strong concentration of positive weights at the far-right edge (largest look-back indices, i.e., the most recent observations), indicating that the trend forecast is dominated by the latest level/slope information. Earlier history contributes comparatively weakly and more uniformly, consistent with a trend estimator that prioritizes the recent trends of the data. Moreover, the weight pattern is also more stable across the forecast steps, implying that the trend head implements a relatively consistent extrapolation mechanism for all horizons, while the seasonal head accounts for more horizon-specific corrections. This can be further attributed to the fact that the intermittent nature of the data during training the model to focus on the more recent look-back index and gradually decrease the attention to the earlier historical time steps to ultimately make the error residual in the loss function for the physics and data components as close as possible to/near $\approx 0$.

For a given forecast step $t$, PI-DLinear forms the prediction as a weighted sum over the look-back window as
\begin{equation}
\hat{y}_t=\sum_{k=0}^{L-1} w_{t,k}\,x_k,
\end{equation}
where we ignore the bias term and the final summation over the components for simplicity. If $w_{t,k} > 0$, then $x_k$ yields a \emph{positive contribution} to $\hat{y}_t$. Holding everything else fixed, a change $\Delta x_k$ induces a change in the forecast of approximately $w_{t,k}\,\Delta x_k$. If $w_{t,k} < 0$, then $x_k$ yields a \emph{negative contribution} to $\hat{y}_t$. Holding everything else fixed, a change $\Delta x_k$ induces a change in the forecast of approximately $w_{t,k}\,\Delta x_k$ (negative when $\Delta x_k>0$).
This provides a local linear interpretation of the heatmap, where each cell indicates how strongly and with what sign a specific look-back index contributes to the prediction at a given horizon.

\end{document}